\documentclass[journal]{IEEEtran}

\ifCLASSINFOpdf
\else
   \usepackage[dvips]{graphicx}
\fi
\usepackage{cite}
\usepackage{url}
\usepackage{balance}
\usepackage{hyperref}
\usepackage{exscale}
\usepackage{relsize}

\usepackage[table]{xcolor}
\usepackage{amsmath}
\usepackage{amssymb}
\usepackage{mathrsfs} %
\usepackage{dsfont}
\usepackage{algorithmic}
\usepackage{algorithm}
\usepackage{graphicx}
\usepackage{subfigure}
\usepackage{multirow}
\usepackage{graphicx}

\begin{document}

\title{TriPINet: Tripartite Progressive Integration Network for Image Manipulation Localization}

\author{Wei-Yun Liang, Jing Xu, and Xiao Jin
\thanks{This work was supported in part by the Science and Technology Planning Project of Tianjin, China (17JCZDJC30700, 18ZXZNGX00310), Tianjin Natural Science Foundation (19JCQNJC00300), and Fundamental Research Funds for the Central Universities of Nankai University (63201192, 63211116). (Wei-Yun Liang and Jing Xu contributed equally to this work.) (Corresponding author: Xiao Jin.) 

Wei-Yun Liang, Jing Xu and Xiao Jin are with the College of Artificial Intelligence, Nankai University, Tianjin 300350, China (e-mail: jinxiao@nankai.edu.cn).

}
}

\markboth{ }
{Shell \MakeLowercase{\textit{et al.}}: Bare Demo of IEEEtran.cls for IEEE Journals}
\maketitle

\begin{abstract}
	Image manipulation localization aims at distinguishing forged regions from the whole test image. Although many outstanding prior arts have been proposed for this task, there are still two issues that need to be further studied: 1) how to fuse diverse types of features with forgery clues; 2) how to progressively integrate multistage features for better localization performance. In this paper, we propose a tripartite progressive integration network (TriPINet) for end-to-end image manipulation localization. First, we extract both visual perception information, e.g., RGB input images, and visual imperceptible features, e.g., frequency and noise traces for forensic feature learning. Second, we develop a guided cross-modality dual-attention (gCMDA) module to fuse different types of forged clues. Third, we design a set of progressive integration squeeze-and-excitation (PI-SE) modules to improve localization performance by appropriately incorporating multiscale features in the decoder. Extensive experiments are conducted to compare our method with state-of-the-art image forensics approaches. The proposed TriPINet obtains competitive results on several benchmark datasets.
\end{abstract}


\IEEEpeerreviewmaketitle

\section{Introduction}

\IEEEPARstart{T}{he} proverb ``\textit{seeing is believing}" refers to that what we can visualize, we can trust and rely on. A foundational assumption of all the image processing techniques is the authenticity of input images. However, with the rapid development of image editing tools, digital images can be easily tampered and spread on websites. Many amateurs without any professional skills can effortlessly edit images using publicly available tools. Digital image forgeries inevitably affect our lives. It is harmless to edit images for entertainment purposes. If image editing is used for military intelligence, celebrity rumors, fake news on the Internet, insurance defrauding crime, plagiarism, and other serious issues, it may cause severe consequences. Thus, the authenticity of image content has always been an important factor for law enforcement officers to consider. Distinguishing manipulated images from authentic ones has attracted increasing attention \cite{Verdoliva2020JSTSP,Farid2019ARVS,Kaissis2020NMI}.

\subsection{Motivation}
\subsubsection{how to fuse different types of manipulation clues} Diverse types of forensic features contribute differently in multimedia tampering detection \cite{Han2021TBIOM,Dong2022TPAMI}. Visual perception information, e.g., RGB inputs, is insufficient to perform manipulation localization, since visual imperceptible features, e.g., frequency and noise traces, contain visually unapparent tampering artifacts \cite{Li2021TNNLS}. Fusion strategies by concatenation or elementwise addition operation can not achieve satisfactory performance in some complex scenes. Thus, how to effectively fuse these features from various domains becomes a significant problem in this task. Till now, most related work use early fusion \cite{Wu2019CVPR,Hu2020ECCV,Rao2021ICCV} or late fusion \cite{Zhou2018CVPR,Dong2022TPAMI}, only few of them use multistage middle fusion. In neural networks with multiple branches, multistage fusion introduces cross-modality interactions into different layers, which can provide additional guidance for feature learning and complementarity between low-level and high-level representations \cite{Ji2021IS}. Consequently, to better explore diverse types of tampering traces, a suitable fusion strategy needs to be further analyzed in this task. 

\subsubsection{how to localize the tampered regions in pixel-level}
Although multimedia forensics has been studied for years, most prior arts focus on image-level authenticity classification. Until recently, only few works analyze tampering localization at pixel-level \cite{Zhuo2022TIFS}. In practical demands, law enforcers should not only classify the fake images, but also justify where the image areas were tampered \cite{Zhuang2021TIFS}. Thus, the localization issue in image forensics needs further study and improvement. Compared with image-level classification, pixel-level prediction is a more precise task in which multiscale features play a significant role. Therefore, how to progressively integrate the multiscale information in the decoder is of great importance in pixel-level localization.

\subsection{Contribution}
The main contributions of this article can be summed up as follows.
\begin{itemize}
	\item In this paper, we present a tripartite progressive integration network (TriPINet), which is an end-to-end triplet architecture designed for image manipulation localization. Both visual perception information, e.g., RGB images, and visual imperceptible features, e.g., frequency and noise traces, are taken into consideration.  
	\item To fuse different types of forensic features, we propose a novel guided cross-modality dual-attention (gCMDA) module. The feature maps obtained by RGB, frequency and noise branches are effectively fused by the gCMDA module to select the most discriminative features for this task.
	\item We also design progressive integration squeeze-and-excitation (PI-SE) modules to localize the tampered areas at the pixel-level. Based on the nature of pooling operations, global max pooling and global average pooling are alternately adopted in the low-level and high-level stages to progressively integrate the multiscale features for better localization results.
	\item Extensive experiments are conducted in several publicly available datasets. The proposed TriPINet achieves competitive performance against state-of-the-art forensic approaches under different evaluation metrics. In addition, ablation studies and robustness analysis are implemented, which demonstrate the validness of the proposed modules in various conditions.
\end{itemize}


\section{Related Work}
\subsection{Image Manipulation Localization}
Most existing image tampering localization methods focus on three semantic forgeries, including splicing, copy-move and removal \cite{Zhuo2022TIFS},\cite{Rao2021SP}. 
Image splicing refers to inserting some regions of donor image into another host image \cite{Rao2021SP}.
Bi et al. \cite{Bi2021ICCV_RTAG} adversarially train two generators based on a multi-decoder-single-task strategy for image splicing detection and localization.
Wang et al. \cite{Wang2022PR} observe image splicing destroys the periodic correlation pattern caused by color filter array (CFA) interpolation. They adopt a linear interpolation model and a covariance matrix to estimate the interpolation coefficients.
Niu et al. \cite{Niu2021TIFS} break the assumptions in previous splicing detection based on double compression artifacts. They propose a general framework to perform image splicing detection, localization and attribution of tampered regions coming from different donor images.

Copy-move operation duplicates the source regions to the target regions within the same image.
Liu et al. \cite{Liu2022TIP} integrate deep matching and keypoint matching in end-to-end manner.
Zhu et al. \cite{Zhu2020TII_AR-Net} propose an adaptive attention and residual refinement network (AR-Net) based on the inconsistence of illumination and contrast. 
Islam et al. \cite{Islam2020CVPR} design a generative adversarial network (GAN) with a dual-order attention model for image copy-move detection and localization. In addition, several approaches are proposed for detecting hybrid types of forgeries \cite{Wang2022CVPR,Kwon2022IJCV,Zhuo2022TIFS,Zhuang2021TIFS,Li2021TNNLS,Rao2021SP,Bappy2019TIP_H-LSTM}.



\subsection{Attention Mechanism}

The attention mechanism aims to focus on the prominent features in computer vision tasks. 
Hu et al. \cite{Hu2018CVPR} propose a squeeze-and-excitation (SE) block to flexibly update channel-wise features. 
Roy et al. \cite{scSE} combine the channel and spatial attention units in parallel, called concurrent spatial and channel squeeze $\&$ excitation (scSE) module. 
Woo et al. \cite{CBAM} design the convolutional block attention module (CBAM), which connects channel and spatial attention in a lightweight manner.
Park et al. \cite{BAM} introduce a bottleneck attention module (BAM) to select salient areas in a 3D attention map. 
Gao et al. \cite{Gao2019CVPR} develop a global second-order pooling (GSoP) block.  

Although attention mechanisms have been utilized for image forensics, some previous approaches only adopted a single type of attention module 
. The effectiveness of the hybrid attention module is not fully explored in this task. Furthermore, most previous methods employ attention mechanisms to select discriminative features from only one type of manipulation clues. How to fuse different types of forensic features with attention modules has received relatively much less concern. Thus, we develop the gCMDA module to fully exploit the collaborative relationship among different types of forensic features. We elaborate this module in Section \ref{sect:gCMDA}. 


\subsection{Progressive Integration}
It is widely acknowledged that the features of different stages are complementary in the neural networks. The deeper features contain global information and are more likely to locate the tampered regions, while the shallower features supply more spatial details. 
Benefiting from progressively integrating multiscale features, multistage networks have been widely used in pixel-level tasks in the computer vision community. 
Li et al. \cite{LAP} gradually integrate multistage feature maps to handle image dehazing in complex scenes. 
Mao et al. \cite{Mao2021TMM} progressively combine the low-level spatial features with the high-level semantic features for salient object detection.
Chen et al. \cite{Chen2018CVPR} observe the complementary information across multiple stages, which is not fully taken into account by previous works. They present a progressively complementarity-aware fusion network for RGB-D salient object detection. Liu et al. \cite{Liu2022TCSVT} leverage multiscale features to produce image manipulation localization masks in a coarse-to-fine fashion. 
These methods have proved that the information in high-level stages can refine the features from low-level stages.

Thus, we adaptively recalibrate the discriminative features from different stages. 
As mentioned in some recent work \cite{Sun2022SPL}, inappropriate pooling operation may destroy some valuable information for image manipulation localization. Therefore, we design the PI-SE module by carefully selecting diverse types of pooling operations in different stages of the decoder. The details of this module
are presented in Section \ref{sect:PISE}.

\begin{figure*}[t]
	\centering
	\includegraphics[width=1.00\textwidth]{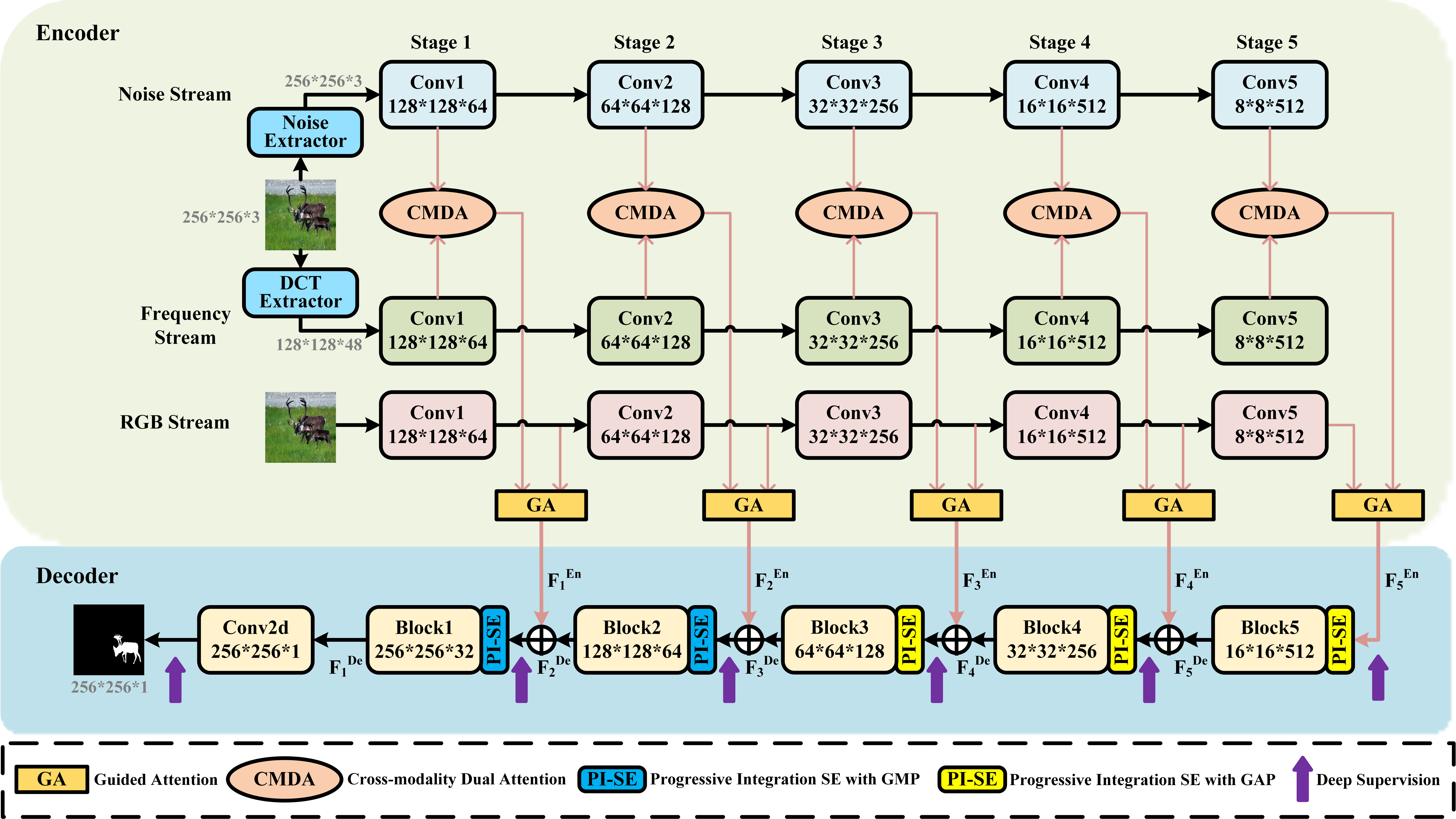}
	\centering
	\caption{\textbf{The overall framework of our proposed TriPINet.} The encoder extracts discriminative features from different domains. The gCMDA module is composed of cross-modality dual attention (CMDA) and guided attention (GA), which is designed to fuse complementary features. The PI-SE modules are proposed to progressively integrate multiscale information in decoder. During training, deep supervision is employed at each stage.   
	}
	\label{fig:FRAMEWORK}
\end{figure*}

\section{Methodology}
\subsection{Overview and Notations}

The architecture of the proposed network is shown in Fig. \ref{fig:FRAMEWORK}. The input of our network is an RGB image $I\in\mathbb{R}^{H\times W\times3}$, where $H$ and $W$ represent the height and the width of the image. The input is first fed into the encoder to generate features in different domains, e.g., RGB, frequency and noise domains. We design the gCMDA module, which consists of cross-modality dual attention and guided attention, to fully explored the complementary information in different domain features. After obtaining these hybrid features, we propose the PI-SE modules to integrate multistage information together for better localization results. During training, deep supervision is adopted in each stage for fast convergence \cite{Lee2015AISTATS}. The hybrid loss \cite{Qin2019CVPR} substitutes for common binary cross-entropy loss \cite{BCE} to predict accurate boundary. The descriptions of some notations and operations are listed in Table \ref{tab:Notations}.

\begin{table}[]
	\centering
	\caption{Notations and operations used in this paper and their descriptions.}
	\label{tab:Notations}
	\begin{tabular}{c|c}
		\hline
		Notations & Descriptions \\ 
		\hline
		${\rm Conv}_{k \times k}(\cdot)$ & $k \times k$ convolutional layers
		\\
		${\rm Up}_{\times 2}(\cdot)$ & upsampling two times by bilinear interpolation
		\\
		${\rm BN}(\cdot)$ & batch normalization (BN) 
		\\
		${\rm ReLU}(\cdot)$ & Rectified Linear Unit (ReLU) activation function 
		\\
		${\rm Softmax}(\cdot, \cdot)$ & elementwise Softmax layer
		\\
		$\sigma(\cdot)$ & Sigmoid activation function 
		\\
		${\rm GAP}(\cdot)$ & global average pooling (GAP) 
		\\
		${\rm GMP}(\cdot)$ & global max pooling (GMP) 
		\\
		${\rm Cat}[\cdot, \cdot]$ & concatenation 
		\\
		${\rm SE}(\cdot)$ & squeeze-and-excitation (SE) attention module
		\\
		${\rm PISE}_i(\cdot)$ & PI-SE attention module in the $i$-th stage of the decoder
		\\ 
		$\otimes$ & elementwise multiplication 
		\\
		$\oplus$ & elementwise addition 
		\\
		\hline
		$i$ & the $i$-th stage in the neural networks
		\\
		$\mathcal{F}_i$ & feature maps outside a module in the $i$-th stage
		\\
		$f_i$ & feature maps inside a module in the $i$-th stage
		\\
		\hline
	\end{tabular}
\end{table}

\subsection{Encoder}\label{sect:encoder}
The proposed TriPINet is a tripartite network.
The encoder of the proposed network consists of three similar branches, namely the RGB stream, the frequency stream, and the noise stream, respectively. 
All the three branches adopt VGG-16 \cite{Simonyan2014VGG} pre-trained on the ImageNet dataset \cite{Russakovsky2015IJCV} as backbones to extract hierarchical features. 
The output feature maps of each backbone in the $i$-th stage are denoted as $\mathcal{F}_i^{RGB}$, $\mathcal{F}_i^{Freq}$ and $\mathcal{F}_i^{Noise}$, $i \in \{1, 2, 3, 4, 5\}$.

\subsubsection{RGB Stream} The RGB stream uses RGB images as inputs. This branch aims to extract visual perceptible tampering artifacts, such as unnatural tampered region boundaries and high contrast differences between tampered and authentic regions \cite{Li2021TNNLS}. If visual imperceptible features incorrectly suppress some content information, the semantic clues in RGB images can provide some complementary guidance \cite{Dong2022TPAMI}. 

\subsubsection{Frequency Stream} 
\begin{figure}[t]
	\centering
	\includegraphics[scale=0.35]{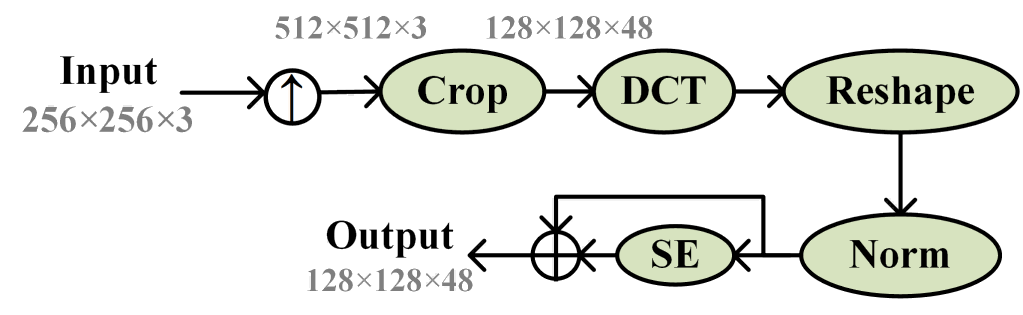}
	\centering
	\caption{\textbf{The architecture of the DCT extractor in frequency stream.} ``DCT" denotes the DCT transformation. ``Norm" represents the channel-wise normalization. ``SE" indicates the squeeze-and-excitation module \cite{Hu2018CVPR}.}
	\label{fig:DCT}
\end{figure}
\begin{figure}[t]
	\centering
	\includegraphics[scale=0.12]{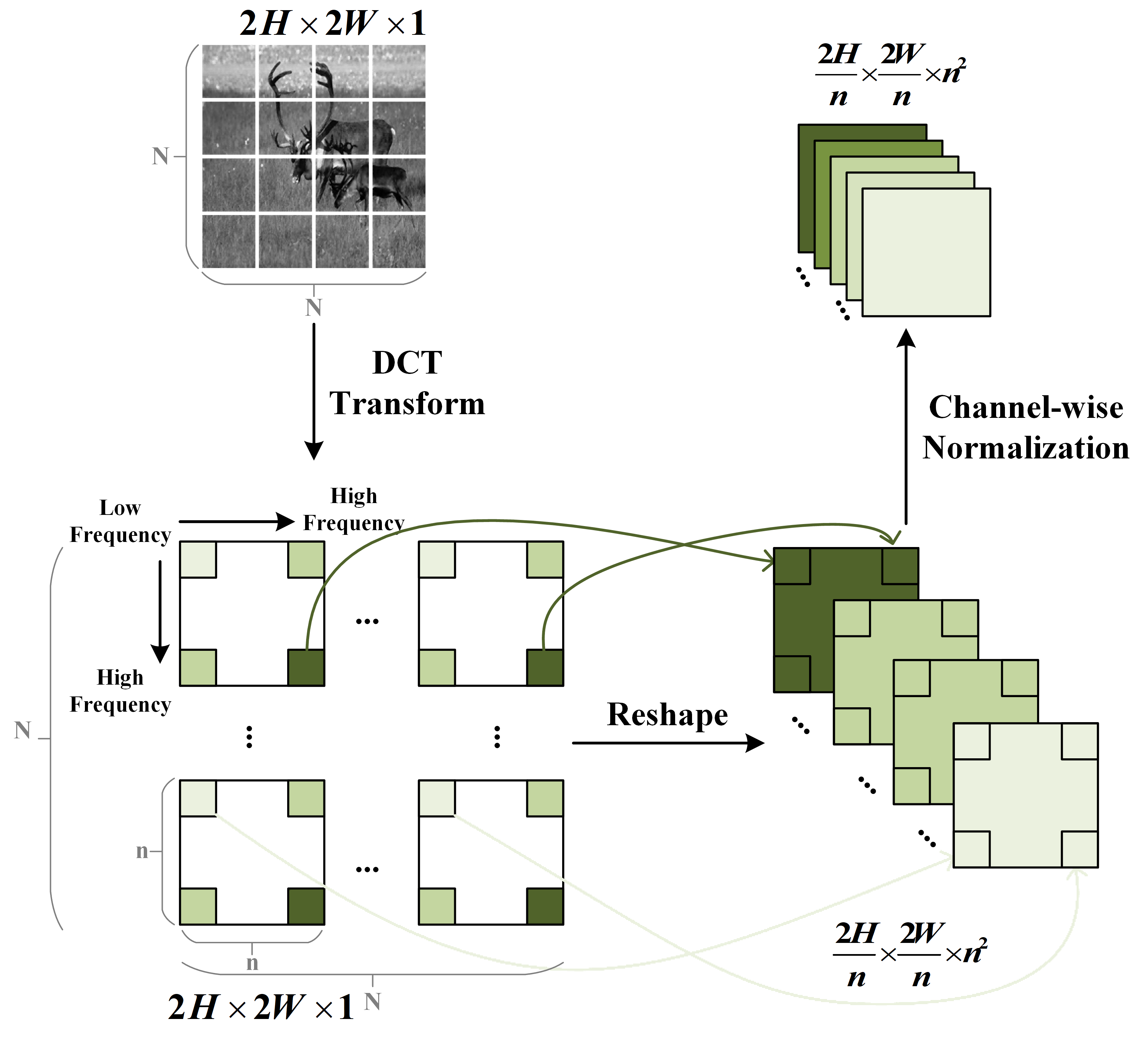}
	\centering
	\caption{\textbf{The illustration of the reshape operation in the DCT extractor.} The input feature maps has three channels, and only one channel is shown here for brevity. Light green represents low frequency features and dark green signifies high frequency components.
	}
	\label{fig:Assemble}
\end{figure}

Frequency domain information may contain key forgery clues. This stream first adopts a DCT extractor to capture forgery traces \cite{Gao2022TKDE_TBNet}. The whole process of the DCT extractor is illustrated in Fig. \ref{fig:DCT}. Given an RGB image $I\in\mathbb{R}^{H\times W\times3}$, it is first converted to the YCbCr color space and upsampled two times to make the shape compatible to the next phase. 
Then, the feature maps are cropped into $N \times N$ patches and discrete cosine transform (DCT) \cite{Ahmed1974DCT} is performed on these patches to obtain $N \times N$ DCT coefficients for each channel. 
Each patch is further cropped into $n \times n$ blocks. 
The blocks that represent the same frequency are assigned in the same channel according to their position in the DCT coefficients. 
Channel-wise normalization are further adopted on these feature maps. The procedure of the reshape operation is shown in Fig. \ref{fig:Assemble}.

To deflect irrelevant frequency, we use a squeeze-and-excitation (SE) module \cite{Hu2018CVPR} to adaptively recalibrate frequency responses. A shortcut connection is also integrated into the SE module to enhance the features.
After that, the DCT extractor outputs the frequency features $I^{DCT}\in\mathbb{R}^{H\times W\times48}$. 
The useful frequency information is enhanced and the irrelevant components are suppressed. 

The first stage of the frequency stream uses a $1 \times 1$ convolutional layer and a batch normalization layer to reshape the feature map to the same size of the second stage. 
Finally, we obtain the output features of the first stage, which is denoted as $\mathcal{F}_1^{Freq}\in\mathbb{R}^{H\times W\times64}$. Except for the first stage, the following part of the frequency stream is the same as the VGG-16 network. 

\subsubsection{Noise Stream} 
\begin{figure}[t]
	\centering
	\includegraphics[scale=0.35]{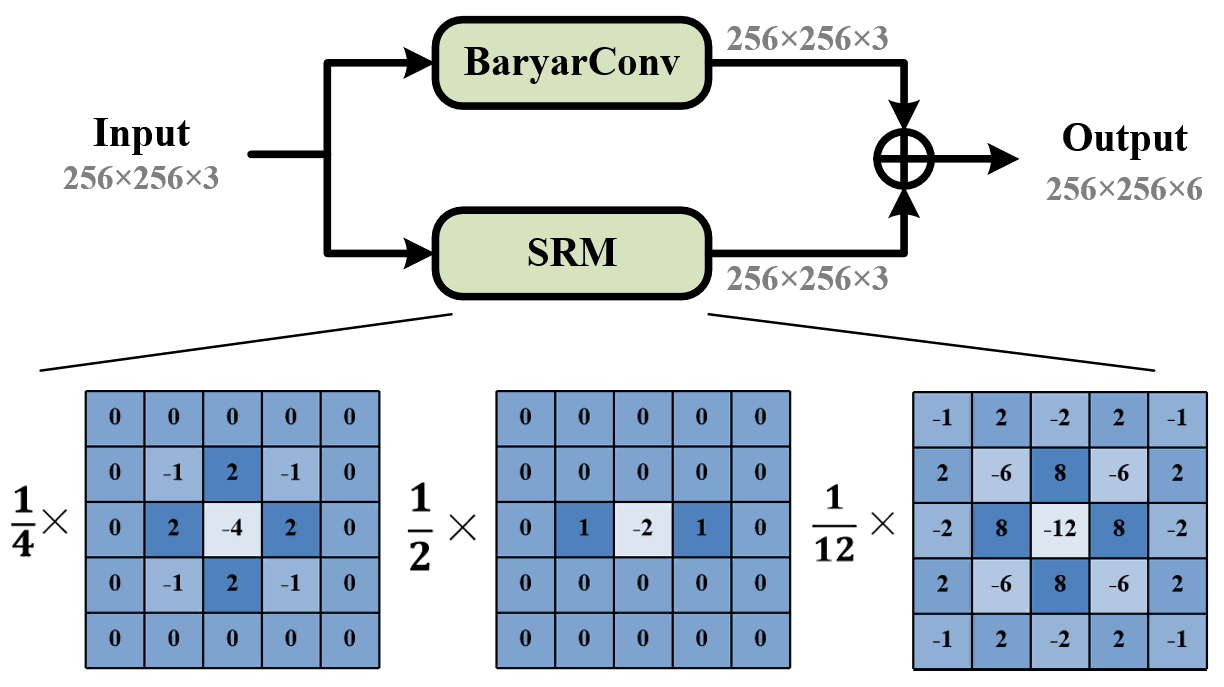}
	\centering
	\caption{\textbf{The architecture of the noise extractor in noise stream.}}
	\label{fig:Noise}
\end{figure}

To suppress the content information, it is essential to exploit the noise view. The noise stream consists of a noise extractor and a VGG-16 backbone. The structure of the noise extractor is shown in Fig. \ref{fig:Noise}. Concretely, the noise extractor adds the output features of BayarConv layer \cite{Bayar2018TIFS} and SRM layer \cite{Fridrich2012TIFS}. The BayarConv layer enhances the noise inconsistency between tampered and original regions. The trainable kernels of the BayarConv layer have two constraints in each training iteration, which are defined as,
\begin{equation}\label{BayarConv}
\begin{cases}
w_h(0,0) = -1, \\
\sum_{m,n \neq 0} w_h(x,y)=1,
\end{cases}
\end{equation}
where $w_h(x,y)$ denotes the parameter in coordinate $(x,y)$ of the $h$-th constrained kernel, and $(0,0)$ denotes the center position. 

The SRM layer has thirty basic fixed convolutional kernels that extract the co-occurrence information of the image residuals. It is proved that only three of the basic SRM kernels have significant effects while applying all thirty kernels brings little performance improvement \cite{Zhou2018CVPR}. Thus, we only use the three effective kernels. The detailed parameters of these three kernels are shown at the bottom of Fig. \ref{fig:Noise}. The kernel size of both BayarConv layer and SRM layer is $5 \times 5 \times 3$, and the number of output channels is 3. The remaining parts of the noise stream are the same as the VGG-16 network. 

\subsection{Guided Cross-Modality Dual-Attention (gCMDA) module}\label{sect:gCMDA}
\begin{figure}[t]
	\centering
	\includegraphics[scale=0.13]{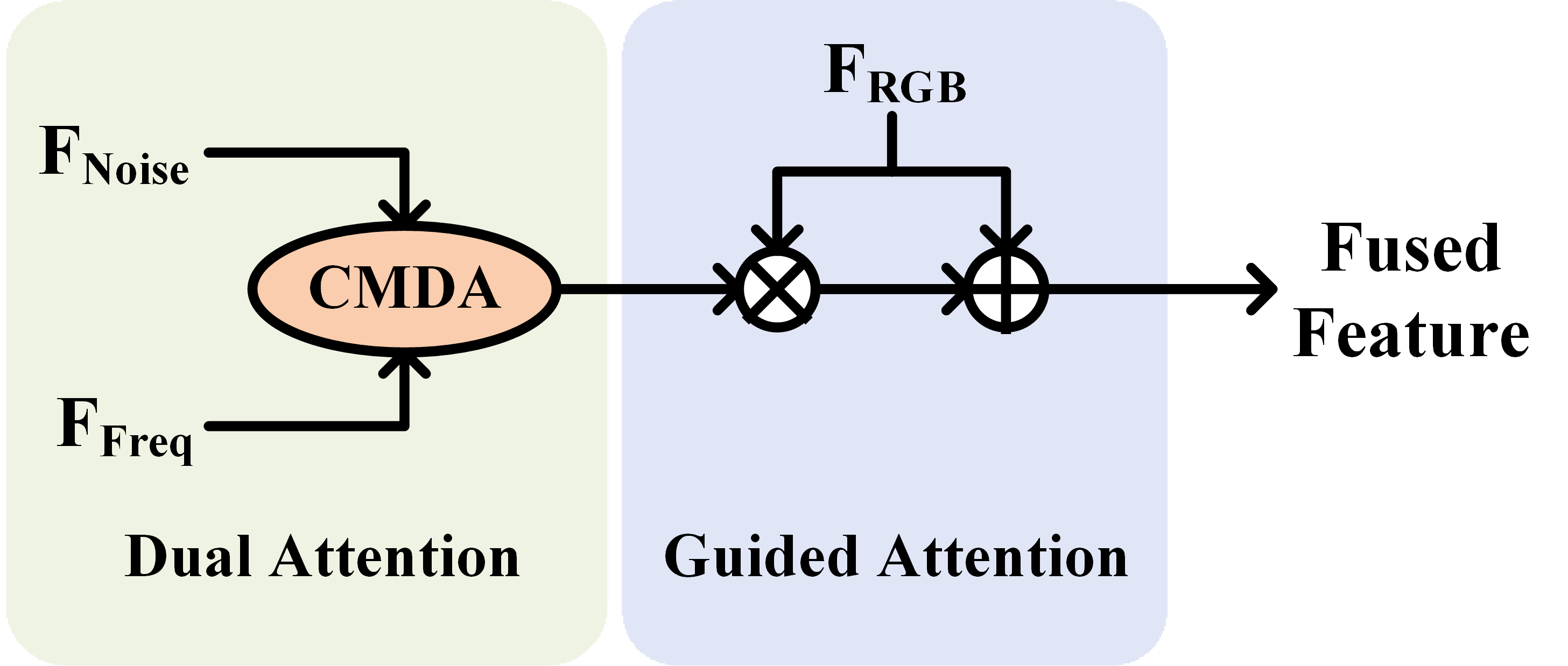}
	\centering
	\caption{\textbf{The structure of the gCMDA module.} First, the noise and frequency features are fused by a CMDA module to obtain visual imperceptible features. Then, the visual imperceptible features and RGB features are further fused by a GA module to generate the outputs of encoder in each stage.}
	\label{fig:gCMDA}
\end{figure}

After obtaining diverse types of features, we select the most discriminative parts for further image manipulation localization. Thus, we propose the gCMDA module. This fusion module consists of two key components, namely cross-modality dual-attention and guided attention. Fig. \ref{fig:gCMDA} illustrates this structure.

\subsubsection{Cross-Modality Dual-Attention}
\begin{figure}[t]
	\centering
	\includegraphics[scale=0.425]{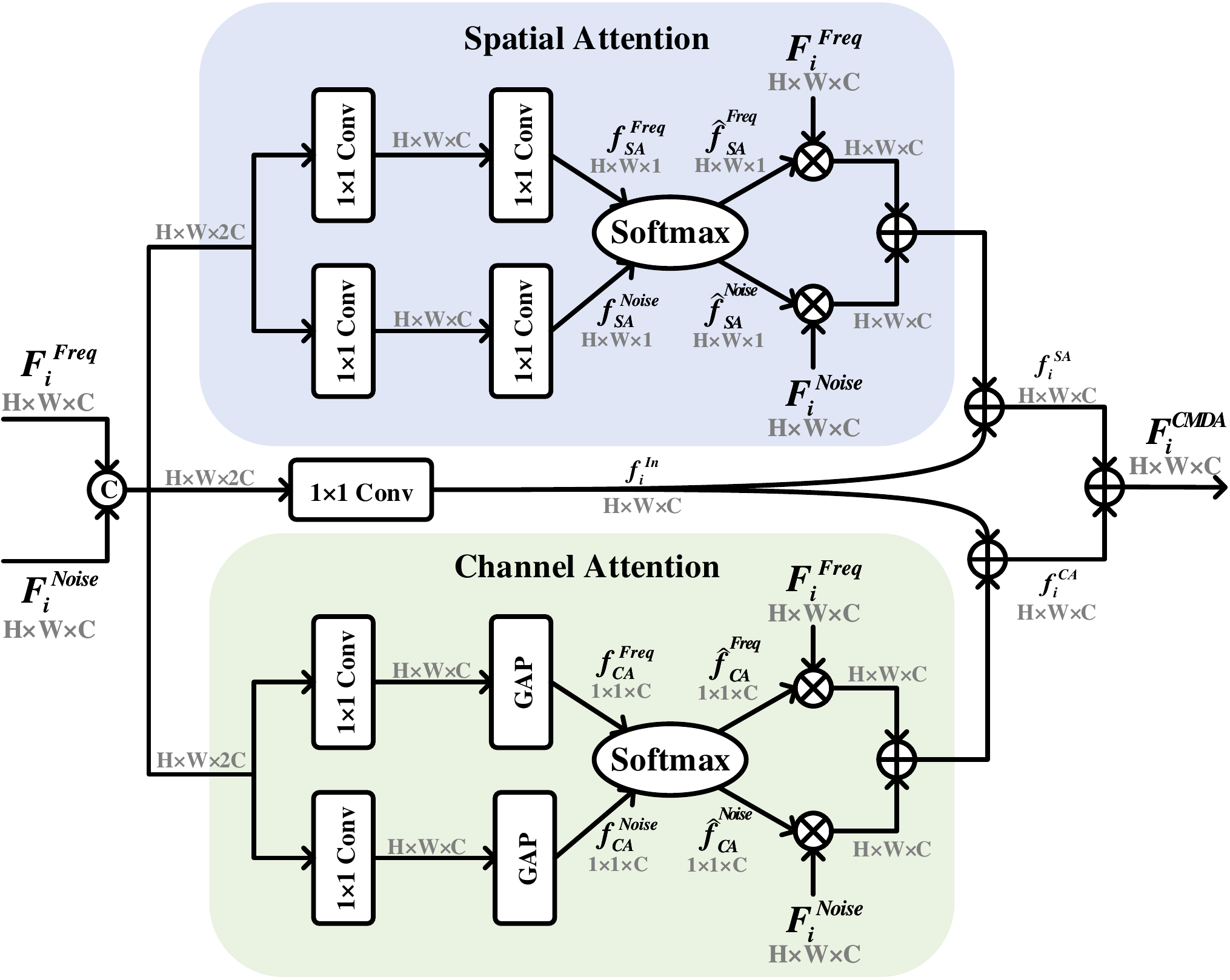}
	\centering
	\caption{\textbf{The structure of the CMDA module.} The CMDA consists of three branches, which are spatial attention branch, intermediante branch and channel attention branch, respectively.}
	\label{fig:CMDA}
\end{figure}

Parallel spatial attention and channel attention are employed to fuse the the visual imperceptible features $ \{ \mathcal{F}_i^{Freq}, \mathcal{F}_i^{Noise} \} \in \mathbb{R}^{H\times W\times C}$ in the cross-modality dual-attention, which is shown in Fig. \ref{fig:CMDA}. 

An intermediate branch between spatial and channel attention is added to enhance the performance. This branch adopts a $1 \times 1$ convolutional layer to reduce the number of channels. The output of this branch is denoted as $f_i^{In} \in \mathbb{R}^{H\times W\times C}$, which can be calculated by the following formula, 
\begin{equation}
f_i^{In} = {\rm Conv}_{1\times1}({\rm Cat}[\mathcal{F}_i^{Freq}, \mathcal{F}_i^{Noise}]),
\end{equation}
where ${\rm Cat}[\cdot,\cdot]$ indicates the concatenation operation, ${\rm Conv}_{1 \times 1}(\cdot)$ represents the $1 \times 1$ convolutional layer.

Spatial attention reweights features to focus on some useful spatial information. In the spatial attention branch, the input features $\mathcal{F}_i^{Freq}$ and $\mathcal{F}_i^{Noise}$ are first concatenated, then the channel number is reduced by two $1 \times 1$ convolutional layers. After that, we use a Softmax layer to obtain the spatial attention weights, which is denoted as $\{ \hat{f}_{SA}^{Freq},\hat{f}_{SA}^{Noise} \} \in \mathbb{R}^{H\times W\times 1}$. This process can be defined as, 
\begin{align}
& f_{SA}^{Freq} = {\rm Conv}_{1\times1}({\rm Conv}_{1\times1}({\rm Cat}[\mathcal{F}_i^{Freq}, \mathcal{F}_i^{Noise}])), \\
& f_{SA}^{Noise} = {\rm Conv}_{1\times1}({\rm Conv}_{1\times1}({\rm Cat}[\mathcal{F}_i^{Freq}, \mathcal{F}_i^{Noise}])), \\
& \{ \hat{f}_{SA}^{Freq}, \hat{f}_{SA}^{Noise} \} = {\rm Softmax}(f_{SA}^{Freq}, f_{SA}^{Noise}),  
\end{align}
where $\{ f_{SA}^{Freq}, f_{SA}^{Noise} \} \in \mathbb{R}^{H\times W\times 1} $ represent intermediate variables. The corresponding elements of $\hat{f}_{SA}^{Freq}$ and $\hat{f}_{SA}^{Noise}$ sum up to 1. The output of the spatial attention branch, denoted as $f_i^{SA} \in \mathbb{R}^{H\times W\times C}$, can be expressed as below,
\begin{equation}
    f_i^{SA} = (\hat{f}_{SA}^{Freq} \otimes \mathcal{F}_i^{Freq} ) \oplus (\hat{f}_{SA}^{Noise} \otimes \mathcal{F}_i^{Noise}) \oplus f_i^{In},
\end{equation}
where $\otimes$ and $\oplus$ indicate elementwise multiplication and addition, respectively.

Channel attention recalibrates channel-wise features to emphasize discriminative channel feature maps. Compared with the spatial attention branch, the channel attention branch uses global average pooling (GAP) layers instead of a $1 \times 1$ convolutional layer to obtain intermediate variables $\{ f_{CA}^{Freq}, f_{CA}^{Noise} \} \in \mathbb{R}^{1\times 1\times C}$. This procedure can be calculated as follows,
\begin{align}
& f_{CA}^{Freq} = {\rm GAP}({\rm Conv}_{1\times1}({\rm Cat}[\mathcal{F}_i^{Freq}, \mathcal{F}_i^{Noise}])), \\
& f_{CA}^{Noise} = {\rm GAP}({\rm Conv}_{1\times1}({\rm Cat}[\mathcal{F}_i^{Freq}, \mathcal{F}_i^{Noise}])), \\
& \{ \hat{f}_{CA}^{Freq}, \hat{f}_{CA}^{Noise} \} = {\rm Softmax}(f_{CA}^{Freq}, f_{CA}^{Noise}), 
\end{align}
where ${\rm GAP}(\cdot)$ is the global average pooling layer, $\{ \hat{f}_{CA}^{Freq}, \hat{f}_{CA}^{Noise} \} \in \mathbb{R}^{1\times 1\times C} $ represent the channel attention weights. Their corresponding elements sum up to 1. Assuming $f_i^{CA} \in \mathbb{R}^{H\times W\times C}$ indicates the outputs of the channel attention branch, it can be defined as, 
\begin{align}
& f_i^{CA} = (\hat{f}_{CA}^{Freq} \otimes \mathcal{F}_i^{Freq}) \oplus (\hat{f}_{CA}^{Noise} \otimes \mathcal{F}_i^{Noise}) \oplus f_i^{In}.
\end{align}

Finally, the output of the CMDA module, denoted as $\mathcal{F}_i^{CMDA} \in \mathbb{R}^{H\times W\times C}$, is computed as,
\begin{equation}
    \mathcal{F}_i^{CMDA} = f_i^{SA} \oplus f_i^{CA}.
\end{equation}
However, visual imperceptible may also suppress some semantic information incorrectly. In the next subsection, we adopt guided attention to handle this case.
 
\subsubsection{Guided Attention} 
Since the goal of the object-based forgery is to tamper objects in the multimedia contents, the semantic information in RGB images is also helpful to localize the tampered areas \cite{Dong2022TPAMI}. It is essential to use complementary information from hybrid views to generate more discriminative features. 
In this paper, we regard RGB features as additional guidance to lead the visual imperceptible features to focus on the most significant forgery clues. Inspired by some recent work \cite{Li2019ICCV,Cao2022CVPR}, we name this structure as guided attention (GA) mechanism in the following. The guided attention is performed with elementwise multiplication and addition, 
\begin{equation}
\begin{split}
\mathcal{F}_i^{En} = (\mathcal{F}_i^{RGB} \otimes \mathcal{F}_i^{CMDA}) \oplus \mathcal{F}_i^{RGB},
 \end{split}
\end{equation}
where $\mathcal{F}_i^{En}$ denotes the output of the guided attention. In this way, the additional RGB features $\mathcal{F}_i^{RGB}$ complements the semantic information that may be incorrectly suppressed by visual imperceptible features $\mathcal{F}_i^{CMDA}$ \cite{Li2019ICCV}. $\mathcal{F}_i^{En}$ is also the output of the encoder in the $i$-th stage. In the next subsection, $\mathcal{F}_i^{En}$ will be fed into the decoder.

\subsection{Progressive Integration SE (PI-SE) module}\label{sect:PISE}

Global average pooling (GAP) and global max pooling (GMP) are commonly used in neural networks. GAP and GMP play different roles in the pooling operation. GAP generates the averaged features of the inputs, while GMP captures the most significant values. As a result, GAP preserves the complete region of the object, whereas GMP pay more attention to the prominent information in the feature maps and might miss out in some detail \cite{Mao2021TMM}. It is important to take advantages of these two pooling strategies in attention modules. 
To recover the feature maps in the decoder, we develop a Progressive Integration SE (PI-SE) module, which uses distinct attention structures in different stages. 

\begin{figure}[t]
	\centering
	\subfigure[PI-SE module in stage 3, 4, 5.]{
		\begin{minipage}[t]{0.5\textwidth}
			\centering
			\includegraphics[scale=0.06]{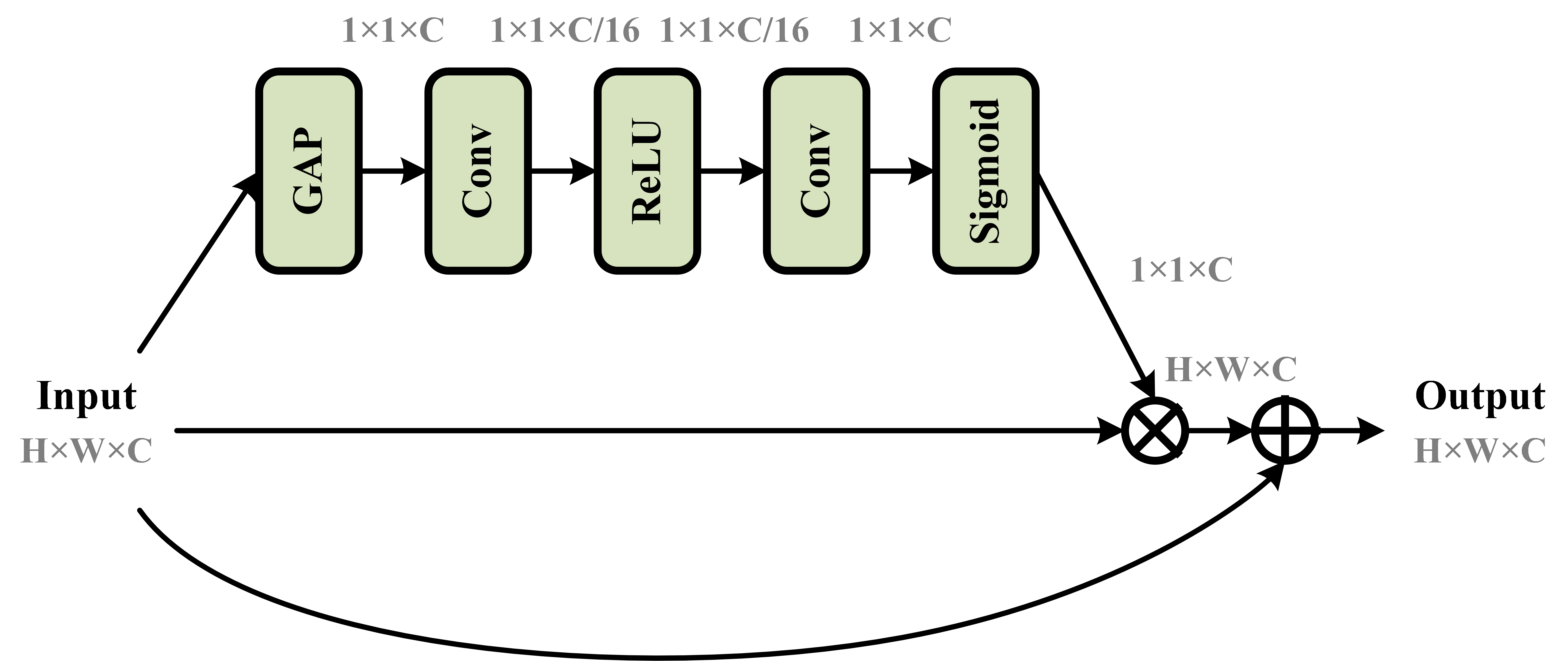}
			\label{fig:PI-SE-GAP}
		\end{minipage}
	}	
	\subfigure[PI-SE module in stage 1, 2.]{
		\begin{minipage}[t]{0.5\textwidth}
			\centering
			\includegraphics[scale=0.06]{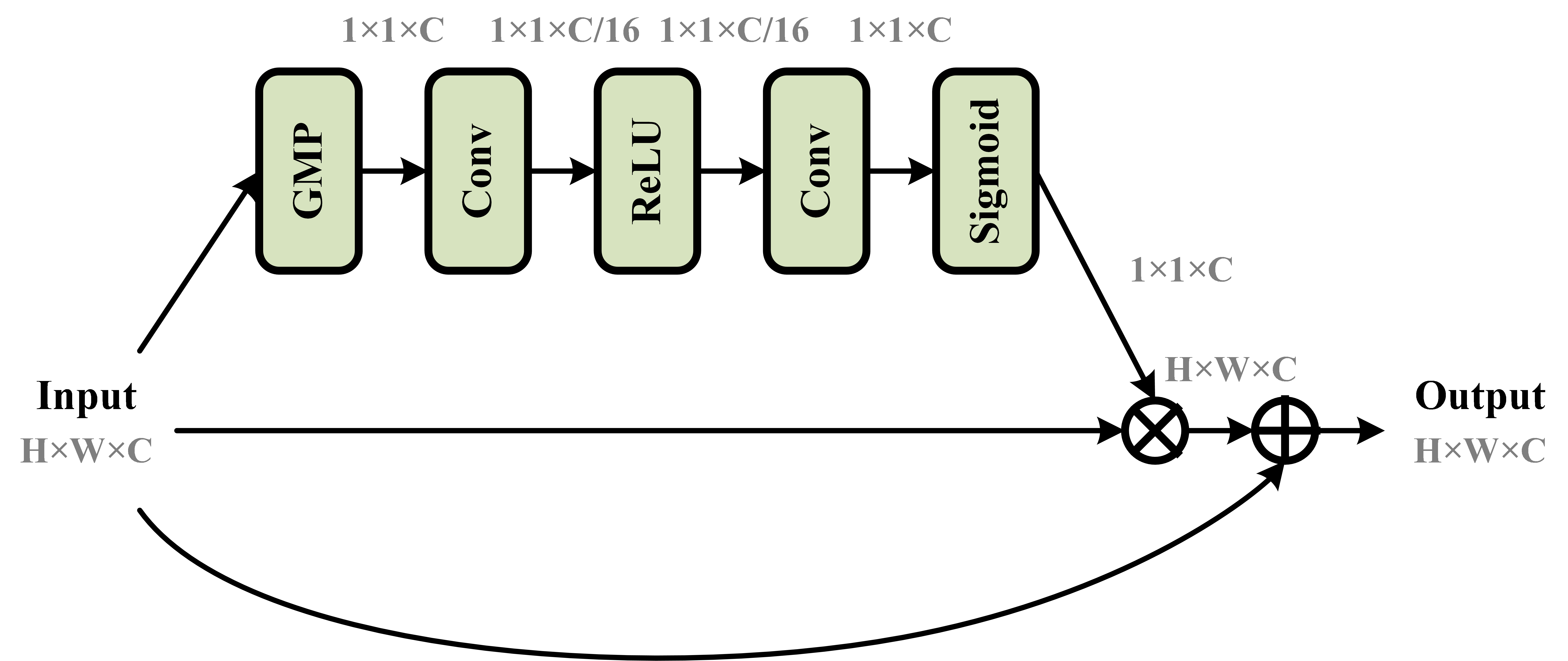}
			\label{fig:PI-SE-GMP}
		\end{minipage}
	}%
	\centering
	\caption{\textbf{The illustration of the PI-SE modules in different stages.} According to the data flow direction in decoder, we use PI-SE module with GAP in the 3rd-5th stages, and PI-SE modules with GMP are adopt in the 1st-2nd stages.}
	\label{fig:PI-SE}
\end{figure}

According to the data flow direction in decoder, we describe the PI-SE from high-level to low-level.
In the third to the fifth stages of the decoder, the resolution of the high-level feature maps is relatively low. Thus, we use PI-SE with GAP to recover the complete manipulated regions. 
Assuming the input of the PI-SE module is denoted as $\mathcal{F}_i^{Fus}$, this calculation process can be expressed by,
\begin{equation}
\begin{split}
f_i^{PISE} = & \sigma({\rm Conv}_{1 \times 1}({\rm ReLU}({\rm Conv}_{1 \times 1}({\rm GAP}(\mathcal{F}_i^{Fus}))))), \\
\mathcal{F}_i^{PISE} = & (f_i^{PISE} \otimes \mathcal{F}_i^{Fus}) \oplus \mathcal{F}_i^{Fus}, i \in \{3,4,5\},
\end{split}
\end{equation}
where $f_i^{PISE} \in \mathbb{R}^{1 \times 1 \times C}$ indicates the attention weights in the PI-SE module and $\mathcal{F}_i^{PISE} \in \mathbb{R}^{H \times W \times C}$ denotes the output features of the PI-SE module. Fig. \ref{fig:PI-SE-GAP} illustrates this case.

In the first two stages of the decoder, the tampering regions have been completely restored by the PI-SE in high-level stages. Therefore, PI-SE with GMP is adopted to focus on the important information and ignore false alarm areas. Meanwhile, the boundaries of the predicted feature maps can be refined in detail \cite{Mao2021TMM}. The PI-SE with GMP can be calculated by,
\begin{equation}
\begin{split}
f_i^{PISE} = & \sigma({\rm Conv}_{1 \times 1}({\rm ReLU}({\rm Conv}_{1 \times 1}({\rm GMP}(\mathcal{F}_i^{Fus}))))), \\
\mathcal{F}_i^{PISE} = & (f_i^{PISE} \otimes \mathcal{F}_i^{Fus}) \oplus \mathcal{F}_i^{Fus}, i \in \{1,2\},
\end{split}
\end{equation}
where ${\rm GMP}(\cdot)$ stands for global max pooling. Fig. \ref{fig:PI-SE-GMP} describes this process intuitively.

\subsection{Decoder}
\label{sect:Decoder}
Decoder aims to gradually increase the resolutions of the feature maps to the raw size of the input images. The decoder is divided into five stages.
Each stage contains a PI-SE module followed by a series of operations.
The input of each PI-SE module in the $i$-th stage involves multiscale information, which is denoted as $\mathcal{F}_{i}^{Fus}$,
\begin{align}
\mathcal{F}_{i}^{Fus} =  \left\{ {\begin{array}{*{20}{l}}
	{\mathcal{F}_{i}^{En} \oplus \mathcal{F}_{i+1}^{De}, i \in \{1, 2, 3, 4\}, }\\
	{\mathcal{F}_{i}^{En},i=5,}
	\end{array}} \right.
\end{align}
where $\mathcal{F}_i^{En}$ is the output of the encoder in the $i$-th stage, $\mathcal{F}_i^{De}$ is the output of the $i$-th block in the decoder.

For brevity, we use ``Block" to represent a set of operations in Fig. \ref{fig:FRAMEWORK}. Each block is composed of a $1 \times 1$ convolutional layer, a batch normalization (BN) layer, a rectified linear unit (ReLU) activation function, and bilinear interpolation for upsampling. The main purpose of these operations is to make the shape compatiable among different stages. 
Then, the output of each stage in decoder can be formulated as,
\begin{align}
\mathcal{F}_i^{De} = {\rm Up_{\times 2}(ReLU(BN}( {\rm Conv}_{1 \times 1}( {\rm PISE}_{i}(\mathcal{F}_{i}^{Fus}))))),
\end{align}
where ${\rm PISE}_i(\cdot)$ denotes the PI-SE module we described in the above subsection. 

After obtaining $\mathcal{F}_1^{De}$, we add a $1 \times 1$ convolutional layer to reduce the channel number to 1. The size of the final feature map output by the decoder is the same as the input images. 

\subsection{Loss Function}\label{sect:Loss}
The proportion of tampered pixels in a manipulated image is relatively small, which leads to a class imbalance problem. Since a large number of negative pixels contribute little to the learning process, the imbalance problem will make the model tends to classify the pixels as authentic \cite{Zhang2021TCSVT,Wu2022TCSVT}. The popular binary cross-entropy (BCE) loss \cite{Shore1980TIT} treats the negative and positive samples equally. Thus, the important information in the positive pixels cannot be fully explored. 

To address this issue, we adopt a hybrid loss that focus on the positive pixels in both tampered boundaries and manipulated regions \cite{Zhu2020TII_AR-Net}\cite{Qin2019CVPR}\cite{Qin2021arXiv}. The hybrid loss consists of the BCE loss, the intersection over union (IoU) loss \cite{Yu2016ACMMM} and the structural similarity (SSIM) loss \cite{Wang2004TIP},
\begin{equation}
\begin{split}
\mathcal{L}_k = \mathcal{L}_k^{BCE} + \mathcal{L}_k^{SSIM} + \mathcal{L}_k^{IoU},
\end{split}
\end{equation}
where $k$ stands for the indexes of the outputs in the deep supervision \cite{Lee2015AISTATS}.

The binary cross-entropy (BCE) treats the foreground and background pixels equally. The BCE loss is defined as,
\begin{equation}
    \mathcal{L}^{BCE} =  - \sum\limits_{j} {[{G_j}\log({P_j}) + (1 - {G_j})\log(1 - {P_j})]},
\end{equation}
where $P$ is the predicted feature maps, $G$ is the groundtruth and $j$ denotes the pixel coordinate. 

The SSIM loss measures the structural similarity between prediction and groundtruth in patch-level. It gives higher weights to boundary pixels that locate between tampered and authentic regions. As a result, the structural information in the boundary areas is retained. The SSIM loss can be expressed by,
\begin{equation}
\mathcal{L}^{SSIM} = 1 - \frac{{(2{\mu _P}{\mu _G} + {\epsilon_1})(2{\sigma _{PG}} + {\epsilon_2})}}{{(\mu _P^2 + \mu _G^2 + {\epsilon_1})(\sigma _P^2 + \sigma _G^2 + {\epsilon_2})}},
\end{equation}
where $\mu _P, \mu _G$ and $\sigma _P, \sigma _G$ are the means and the standard deviations of two patches from the prediction $P$ and the groundtruth $G$, respectively. $\sigma _{PG}$ is the covariance matrix. ${\epsilon_1} = 0.01^2$ and ${\epsilon_2} = 0.03^2$ are constant numbers to prevent division by zero. 

The IoU loss, which is often used in object detection and segmentation, emphasizes more on the large foreground regions. It is formulated as,
\begin{equation}
\mathcal{L}^{IoU} = 1 - \sum\limits_j {\frac{{{P_j}{G_j}}}{{{P_j} + {G_j} - {P_j}{G_j}}}}.
\end{equation}

The outputs of the decoder are used for the final manipulation localization. Besides the final outputs, we also add deep supervision \cite{Lee2015AISTATS} in each stage of the decoder. In this way, our network is supervised with six outputs. Finally, the training loss is defined as the summation of the hybird loss for all the outputs,
\begin{equation}
\mathcal{L} = \sum_{k=1}^6 \mathcal{L}_k,
\end{equation}
where $k \in \{1, 2, 3, 4, 5, 6\} $ denotes the $k$-th supervised output.
In the decoder, the predicted feature maps will be upsampled by bilinear interpolations to make the resolution consistent with the groundtruth.

\section{Experiments}
\subsection{Experimental Setup}
\subsubsection{Datasets} \label{sect:datasets}  
We conduct experiments on seven public benchmark datasets to verify the effectiveness of our proposed method. \textbf{CASIA} \cite{Dong2013CASIA} is composed of two versions. 
CASIAv1 has 461 splicing and 459 copy-move images, and the tampered images are not post-processed. CASIAv2 has 1,828 splicing, 3,295 copy-move and 7,491 authentic images. Post-processing operations such as filtering and blurring are applied to the tampered images. 
\textbf{COLUMB} \cite{Ng2009COLUMB} contains 184 authentic and 180 splicing images. The source tampered images are not post-processed. 
\textbf{Carvalho} \cite{De2013Carvalho} has 100 splicing and 100 authentic images. The manipulated contents are all people. It is post-processed by adjusting contrast and illumination. 

The above three datasets are used for splicing localization, while the following four datasets are adopted for copy-move localization. 
\textbf{Synthetic Dataset} \cite{Bappy2019TIP_H-LSTM} includes 6,200 image copy-move examples, which is 
subjected to resizing and rotation in different factors. 
\textbf{CASIA-CMFD} is produced by \cite{Wu2018ECCV_BusterNet}. It is a subset of the CASIAv2 dataset. There are 1,313 copy-move images and their authentic counterparts. 
\textbf{CoMoFoD} \cite{Tralic2013CoMoFoD} contains 200 source copy-move images. Each source tampered image will undergo 25 post-processing/attacks to produce another 25 tampered images. 
\textbf{COVERAGE} \cite{Wen2016COVERAGE} is a challenge copy-move dataset, which contains 100 copy-move images and their corresponding ground-truth masks. 

\subsubsection{Evaluation Metrics}
Image manipulation localization is a pixel-level binary classification problem, which infers whether each pixel is tampered or not. Following previous works \cite{Zhang2021TCSVT,Wu2018ECCV_BusterNet}, we use F1-score and Intersection over Union (IoU) to quantify the performance. These metrics can be calculated by,
\begin{equation}
    F1 = \frac{2TP}{2TP+FN+FP},
\end{equation}
\begin{equation}
    IoU = \frac{TP}{TP+FN+FP},
\end{equation}
where $TP$ represents the number of correctly classified manipulated pixels; $FN$ denotes the number of misclassified tampered pixels; $TN$ indicates the number of correctly classified authentic pixels; $FP$ is the number of wrongly classified original pixels. 

\begin{table}[t]
	\centering
	\renewcommand\arraystretch{1.5} 
	\caption{Data split protocol for image splicing and copy-move localization, respectively.}
	\label{tab:DataSplit_SP_pix}
	\begin{tabular}{cccc}
		\hline
		Dataset  & Train & Validation & Test \\ \hline
		CASIAv2 \cite{Dong2013CASIA} & 4500  & 122        & 501  \\
		CASIAv1 \cite{Dong2013CASIA} & 341   & 20         & 100  \\
		COLUMB \cite{Ng2009COLUMB}  & 125   & 5          & 50   \\
		Carvalho \cite{De2013Carvalho} & 70    & 10         & 20   \\ \hline
		Synthetic Dataset \cite{Bappy2019TIP_H-LSTM} & 4960 & 620 & 620 \\
		CASIA-CMFD \cite{Wu2018ECCV_BusterNet} & 1051  & 131        & 131  \\
		CoMoFoD  \cite{Tralic2013CoMoFoD}   & 180   & 20         & 20   \\
		COVERAGE \cite{Wen2016COVERAGE}   & 80    & 10         & 10   \\ \hline
	\end{tabular}
\end{table}

\subsubsection{Comparative Methods}
Since the performance of the methods based on deep learning has outperforms the conventional approaches by a large margin, we only use the frameworks based on deep learning for comparison. 
In image splicing localization, 11 state-of-the-art frameworks are included.
\begin{itemize}
	\item \textbf{DenseFCN} \cite{Zhuang2021TIFS} is a fully convolutional network, which adopts dense connections and dilated convolutions for better localization.  
	\item \textbf{MFCN} \cite{Salloum2018JVCIR_MFCN} is a multitask dual-stream network, which mines traces in manipulated region and forged boundaries, respectively.
	\item \textbf{J-LSTM} \cite{Bappy2017ICCV_J-LSTM} combines LSTM with CNN to learn the boundary discrepancy.
	\item \textbf{H-LSTM} \cite{Bappy2019TIP_H-LSTM} introduces resampling features in LSTM to discover the transition between tampered and authentic patches.
	\item \textbf{SG-CRF} \cite{Cun2018ECCVW} leverages local and global features together to obtain better localization performance.
	\item \textbf{ManTra-Net} \cite{Wu2019CVPR} formulates the tampering localization problem as a local anomaly detection problem. It adopts self-supervised learning to learn robust tampered traces. 
	\item \textbf{RRU-Net} \cite{Bi2019CVPRW_RRU-Net} is a ringed residual U-Net for image forgery detection.
	\item \textbf{MTSE-Net} \cite{Zhang2021TCSVT} incorporates multitask learning with squeeze-and-excitation attention for splicing localization.
	\item \textbf{C2R-Net} \cite{Xiao2020IS_C2RNet} cascades a coarse convolutional neural netowrk and a refined CNN.
	\item \textbf{RTAG} \cite{Bi2021ICCV_RTAG} uses a generative adversarial network framework for splicing localization.
	\item \textbf{MCNL-Net} \cite{Wei2020TOMM} employs U-Net controlled by multiple scales to localize the tampered regions.
\end{itemize}
In copy-move localization, two popular methods are involved for comparison.
\begin{itemize}
	\item \textbf{BusterNet} \cite{Wu2018ECCV_BusterNet} uses the Simi-Det branch to detect copy-move regions. Following \cite{Zhu2020TII_AR-Net,Chen2021TMM_CMSDNet,Liu2022TIP}, we compare with the similar branch of BusterNet.\footnote{https://github.com/ntdat017/BusterNet\_pytorch}
	\item \textbf{CMSDNet} \cite{Chen2021TMM_CMSDNet} adopts double-level self-correlation and dilated convolutions to extract high-resolution features.\footnote{https://github.com/imagecbj/A-serial-image-copy-move-forgery-localization-scheme-with-source-target-distinguishment}
\end{itemize}

\subsection{Training and Implementation Details}
Our models are implemented in PyTorch and trained on an NVIDIA RTX 3080Ti GPU. All the input images are resized to $256 \times 256$. We perform data augmentation, such as random horizontal and vertical flipping, on the training data. The VGG-16 backbones are initialized with the weights that are pre-trained on the ImageNet dataset \cite{Russakovsky2015IJCV}. The proposed network is trained end-to-end using the Adam optimizer with a weight decay of 5e-4. The learning rate is initialize to 1e-3 and reduced to half if the validation loss does not drop for five epochs. The batch size is set as 16. We train our model for 100 epochs. The model that achieves the best performance on validation set is chose as the final model to evaluate the performance on test set. Following the same protocol in \cite{Zhou2018CVPR,Salloum2018JVCIR_MFCN,Bappy2019TIP_H-LSTM}, we vary different thresholds and use the highest metric score as the final score.

\subsubsection{Image Splicing Localization}
In this task, we follow the exact same experimental protocol defined in \cite{Zhang2021TCSVT}. We randomly split the data in CASIAv2, CASIAv1, COLUMB and Carvalho datasets for evaluation. The details of the split ratios are shown in Table. \ref{tab:DataSplit_SP_pix}. To address the overfitting caused by data deficiency, we pre-trained our model on CASIAv2. Then, we further fine-tuned our model on the other three small datasets. Note that fine-tuning is a widely-used protocol in image splicing localization \cite{Zhou2018CVPR,Hu2020ECCV}.

\subsubsection{Image Copy-Move Localization}
In image copy-move localization, the source and target pixels are both labeled as forged \cite{Wu2018ECCV_BusterNet,Zhu2020TII_AR-Net,Chen2021TMM_CMSDNet,Liu2022TIP}. 
Previous methods often use a large self-made dataset for pre-training, and evaluate without fine-tuning.
Since these self-made datasets are either private or no longer publicly accessible, we use a subset of the Synthetic Dataset \cite{Bappy2019TIP_H-LSTM} for pre-training.\footnote{https://www.dropbox.com/sh/palus3sq4zvdky0/AACu3s7KA5Fhr\_\\BJUeDOxnTLa?dl=0} 
In the pre-training phase, we randomly divided the Synthetic Dataset into the training, validation and test sets according to the ratio of 8:1:1, which is suggested in \cite{Wu2018ECCV_BusterNet}. 
Then, the pre-trained models are evaluated on CASIA-CMFD, CoMoFoD, COVERAGE, and remaining samples in the Synthetic Dataset to verify the generalization ability. 
Additionally, we also provide performance comparison after fine-tuning on CASIA-CMFD, CoMoFoD, and COVERAGE datasets. The dataset partition is shown in Table. \ref{tab:DataSplit_SP_pix}. Fine-tuning is a common protocol in image splicing localization, while it is neglected by some previous copy-move detection methods.

\begin{table*}[]
\centering
\renewcommand\arraystretch{1.5} 
\caption{\textbf{Performance on pixel-level image splicing localization.} Best result per test set is highlighted in bold. All methods are evaluated following the same protocol defined in \cite{Zhang2021TCSVT} except for those marked with asterisks (*). }
\label{tab:SP_pix}
\begin{tabular}{ccccccccccccccc}
\hline
\multirow{2}{*}{Method} 
& \multicolumn{2}{c}{CASIAv2 \cite{Dong2013CASIA}} & 
& \multicolumn{2}{c}{CASIAv1 \cite{Dong2013CASIA}} & 
& \multicolumn{2}{c}{COLUMB \cite{Ng2009COLUMB}} & 
& \multicolumn{2}{c}{Carvalho \cite{De2013Carvalho}} & 
& \multicolumn{2}{c}{Average} \\ \cline{2-3} \cline{5-6} \cline{8-9} \cline{11-12} \cline{14-15} 
                                                & F1    & IOU   & & F1    & IOU   & & F1    & IOU   & & F1    & IOU   & & F1    & IOU   \\ \hline
J-LSTM (2017ICCV) \cite{Bappy2017ICCV_J-LSTM}     & 0.167 & 0.106 & & 0.431 & 0.309 & & 0.275 & 0.154 & & 0.278 & 0.169 & & 0.288 & 0.185 \\
SG-CRF (2018ECCV) \cite{Cun2018ECCVW}             & 0.145 & 0.091 & & 0.166 & 0.098 & & 0.256 & 0.172 & & 0.234 & 0.136 & & 0.200 & 0.124 \\
MFCN (2018JVCIR) \cite{Salloum2018JVCIR_MFCN}     & 0.376 & 0.304 & & 0.423 & 0.340 & & 0.622 & 0.529 & & 0.321 & 0.215 & & 0.436 & 0.347 \\
H-LSTM (2019TIP) \cite{Bappy2019TIP_H-LSTM}       & 0.194 & 0.134 & & 0.269 & 0.186 & & 0.270 & 0.175 & & 0.161 & 0.093 & & 0.224 & 0.147 \\
ManTra-Net (2019CVPR) \cite{Wu2019CVPR}& 0.201 & 0.126 & & 0.226 & 0.139 & & 0.472 & 0.328 & & 0.325 & 0.202 & & 0.306 & 0.199 \\
RRU-Net (2019CVPRW) \cite{Bi2019CVPRW_RRU-Net}    & 0.533 & 0.475 & & 0.481 & 0.426 & & 0.731 & 0.669 & & 0.247 & 0.180 & & 0.498 & 0.438 \\
DenseFCN (2021TIFS) \cite{Zhuang2021TIFS}         & 0.098 & 0.069 & & 0.174 & 0.125 & & 0.323 & 0.231 & & 0.271 & 0.164 & & 0.217 & 0.147 \\
MTSE-Net (2022TCSVT) \cite{Zhang2021TCSVT}        & 0.765 & 0.714 & & 0.667 & 0.592 & & 0.910 & 0.867 & & 0.353 & 0.251 & & 0.674 & 0.606 \\ \hline
C2R-Net* (2020IS) \cite{Xiao2020IS_C2RNet}        & 0.676 & -     & & -     & -     & & 0.695 & -     & & -     & -     & & -     & -     \\
MCNL-Net* (2020TOMM) \cite{Wei2020TOMM}           & 0.866 & -     & & -     & -     & & 0.772 & -     & & -     & -     & & -     & -     \\ 
RTAG* (2021ICCV) \cite{Bi2021ICCV_RTAG}           & 0.815 & -     & & -     & -     & & 0.823 & -     & & -     & -     & & -     & -     \\ \hline
Ours+VGG11                                      & 0.862 & 0.758 & & 0.809 & 0.679 & & 0.948 & 0.901 & & 0.551 & 0.380 & & 0.793 & 0.680 \\
\textbf{Ours+VGG16} &
  \textbf{0.904} &
  \textbf{0.824} &
  & 
  \textbf{0.812} &
  \textbf{0.683} &
  & 
  \textbf{0.962} &
  \textbf{0.927} &
  & 
  \textbf{0.566} &
  \textbf{0.395} &
  & 
  \textbf{0.811} &
  \textbf{0.707} \\ \hline
\end{tabular}
\end{table*}

\begin{figure*}[t]
	\centering
	\includegraphics[scale=0.11]{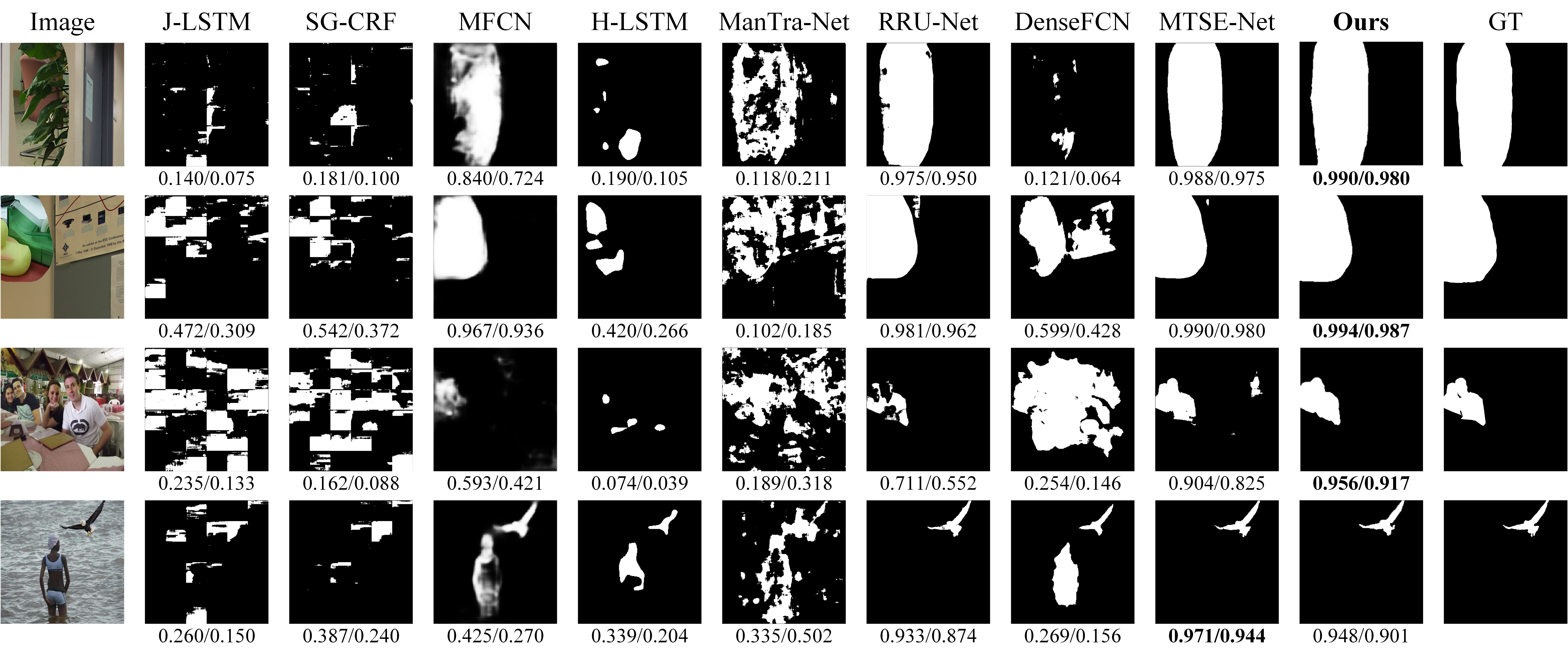}
	\centering
	\caption{Splicing localization results of different methods (F1-score/IoU).}
	\label{fig:SPFig}
\end{figure*}

\subsection{Comparison on Image Splicing Localization}
Quantitive comparison on splicing localization is listed in Table. \ref{tab:SP_pix}. Since we follow the exactly same protocol with \cite{Zhang2021TCSVT}, the evaluation results of other methods in Table \ref{tab:SP_pix} are reported from \cite{Zhang2021TCSVT}.
There are three other approaches using the same datasets with different data split ratios \cite{Xiao2020IS_C2RNet,Wei2020TOMM,Bi2021ICCV_RTAG}. We mark these three with an asterisk (*).
Compared the splicing localization performance F1-score/IoU with MTSE-Net, it can be observed that our method is 13.9\%/11.0\% higher on CASIAv2, 14.5\%/9.1\% higher on CASIAv1, 5.2\%/6.0\% higher on COLUMB and 21.3\%/14.4\% higher on Carvalho. 
The ``Average" represents the average performance over the four standard datsets.
The proposed TriPINet achieves an average F1-score/IoU score of 81.1\%/70.7\%, which is 13.7\%/10.1\% higher than the second-best method, e.g., MTSE-Net \cite{Zhang2021TCSVT}. 
Our model achieves competitive performance against other existing splicing localization methods on the four standard datasets. 
The reason can be explained as follows. Different from other methods only use simple fusion strategies, we design the gCMDA module to fully explore the complementary information in visual perceptible and imperceptible features. In addition, we introduce PI-SE modules to gradually refine the feature maps by multiscale information.

More intuitively, Fig. \ref{fig:SPFig} shows the qualitative localization results. The localization maps of J-LSTM \cite{Bappy2017ICCV_J-LSTM} and SG-CRF \cite{Cun2018ECCVW} shows apparent block artifacts because both of these two methods train images in a patch manner. 
It is also difficult for other mentioned approaches to detect complete outlines of the manipulated regions. 
MTSE-Net \cite{Zhang2021TCSVT} obtains relatively good performance. 
However, it still wrongly localize extra objects as spliced regions, as shown in the second and the third rows of Fig. \ref{fig:SPFig}. 
In the first three rows, the proposed TriPINet can precisely localize the tampered regions.
In the fourth example, although the quantitative metrics of our proposed method are slightly lower than MTSE-Net, the qualitative results are still comparable.

\begin{table*}[t]
\centering
\renewcommand\arraystretch{1.5} 
\caption{\textbf{Performance of pixel-level image copy-move localization of different pre-trained models.} The models are trained on Synthetic Dataset \cite{Bappy2019TIP_H-LSTM}. Best result per test set is highlighted in bold.}
\label{tab:CM_Pretrain}
\begin{tabular}{ccccccccccccccccccc}
\hline
\multirow{2}{*}{Method} 
& \multicolumn{2}{c}{Synthetic \cite{Bappy2019TIP_H-LSTM}} & 
& \multicolumn{2}{c}{CASIA-CMFD \cite{Wu2018ECCV_BusterNet}} & 
& \multicolumn{2}{c}{CoMoFoD \cite{Tralic2013CoMoFoD}} & 
& \multicolumn{2}{c}{COVERAGE \cite{Wen2016COVERAGE}} & 
& \multicolumn{2}{c}{Average} \\ \cline{2-3} \cline{5-6} \cline{8-9} \cline{11-12} \cline{14-15}
                                                & F1    & IoU   & & F1    & IoU     & & F1    & IoU   & & F1    & IoU   & & F1    & IoU\\ \hline
BusterNet (2018ECCV) \cite{Wu2018ECCV_BusterNet}  & 0.872 & 0.772 & & 0.240 & 0.136   & & 0.189 & 0.105 & & 0.510 & 0.342 & & 0.453 & 0.339\\
CMSDNet (2021TMM) \cite{Chen2021TMM_CMSDNet}      & 0.932 & 0.872 & & \textbf{0.285} 
                                                                  & \textbf{0.166}  & & 0.215 & 0.120 & & 0.565 & 0.393 & & 0.499 & 0.388\\
\textbf{Ours}   & \textbf{0.963}    & \textbf{0.929}    & 
                & 0.276             & 0.160             & 
                & \textbf{0.246}    & \textbf{0.140}    & 
                & \textbf{0.622}    & \textbf{0.451}    & 
                & \textbf{0.527}    & \textbf{0.420} \\ \hline
\end{tabular}
\end{table*}

\begin{table*}[t]
\centering
\renewcommand\arraystretch{1.5} 
\caption{\textbf{Performance of pixel-level image copy-move localization using fine-tuned models.} The models are pre-trained on Synthetic Dataset \cite{Bappy2019TIP_H-LSTM} and fine-tuned on each dataset. Best result per test set is highlighted in bold.}
\label{tab:CM_FineTune} 
\begin{tabular}{cccccccccccc}
\hline
\multirow{2}{*}{Method} 
& \multicolumn{2}{c}{CASIA-CMFD \cite{Wu2018ECCV_BusterNet}} & 
& \multicolumn{2}{c}{CoMoFoD \cite{Tralic2013CoMoFoD}} & 
& \multicolumn{2}{c}{COVERAGE \cite{Wen2016COVERAGE}} & 
& \multicolumn{2}{c}{Average} \\ \cline{2-3} \cline{5-6} \cline{8-9} \cline{11-12}
            & F1    & IoU   & & F1    & IoU   & & F1    & IoU   & & F1    & IoU \\ \hline
BusterNet (2018ECCV) \cite{Wu2018ECCV_BusterNet}  & 0.371 & 0.228 & & 0.823 & 0.700 & & 0.677 & 0.512 & & 0.624 & 0.480\\
CMSDNet (2021TMM) \cite{Chen2021TMM_CMSDNet}    & 0.446 & 0.287 & & 0.904 & 0.825 & & 0.706 & 0.546 & & 0.685 & 0.553 \\
\textbf{Ours} & \textbf{0.472} & \textbf{0.309} & 
              & \textbf{0.958} & \textbf{0.919} & 
              & \textbf{0.751} & \textbf{0.602} & 
              & \textbf{0.727} & \textbf{0.610} \\ \hline
\end{tabular}
\end{table*}

\begin{figure}[t]
	\centering
	\includegraphics[scale=0.1]{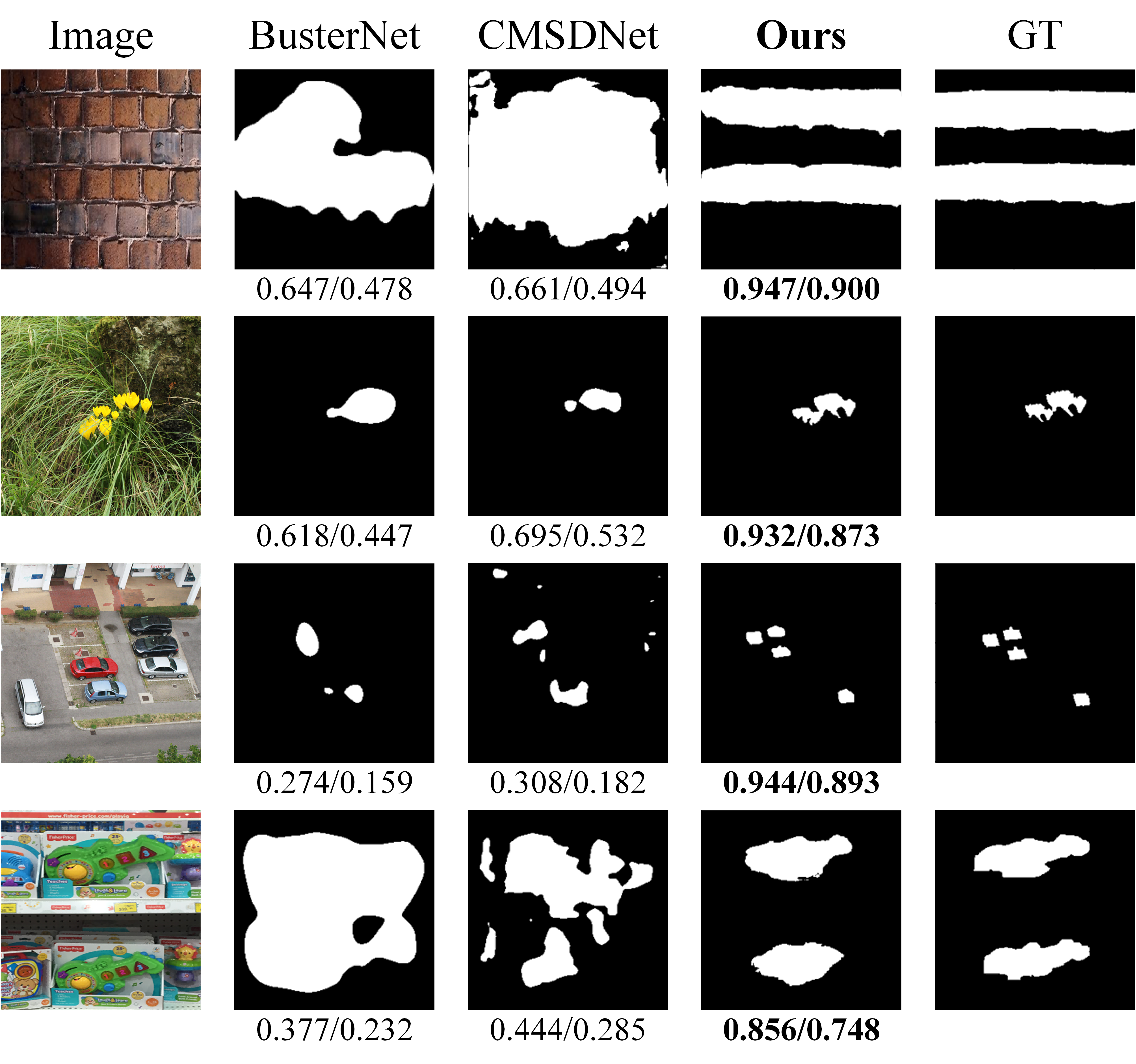}
	\centering
	\caption{Copy-move localization results of different methods (F1-score/IoU).}
	\label{fig:CMFig}
\end{figure}

\subsection{Comparison on Image Copy-Move Localization}
In this subsection, we first pre-trained the proposed TriPINet and other compared methods on the Synthetic Dataset \cite{Bappy2019TIP_H-LSTM}. Then, these pre-trained models are tested on CASIA-CMFD, CoMoFoD, COVERAGE, and remaining samples in the Synthetic Dataset.
As shown in Table. \ref{tab:CM_Pretrain}, our method outperforms other approaches on Synthetic Dataset \cite{Bappy2019TIP_H-LSTM}, CoMoFoD and COVERAGE. 
On the CASIA-CMFD dataset, the F1-score/IoU of our method is slightly lower than CMSDNet by 0.9\%/0.6\%. 
We try to explain the reason for these observations as follows. 
The CASIA-CMFD dataset is a challenging dataset, whose data distribution is very different from Synthetic Dataset \cite{Bappy2019TIP_H-LSTM}.
Additionally, the copy-move detection schemes usually adopt self-correlation modules to match patch-like features \cite{Wu2018ECCV_BusterNet,Zhu2020TII_AR-Net,Chen2021TMM_CMSDNet,Liu2022TIP}. Without such a highly customized module for copy-move localization, our performance will inevitably degrade. However, the proposed TriPINet still achieves comparable results on the average of the four datasets.


After that, we fine-tuned these pre-trained models on CASIA-CMFD, CoMoFoD, and COVERAGE. As shown in Table. \ref{tab:CM_FineTune}, after fine-tuning, our method outperforms existing copy-move localization methods on these three datasets. 
Although not equipped with self-correlation modules, our model can still achieve competitive performance for copy-move localization.
Usually, the pre-training datasets for copy-move detection contain one hundred thousand samples \cite{Wu2018ECCV_BusterNet}. The annotation and training of such a large dataset are labor and time consuming. The Synthetic Dataset we used only has 6,200 samples, nearly 6\% of some large datasets. Fine-tuning can boost performance even without large-scale pre-training, which speeds up the training process.


The qualitative results of copy-move localization methods are shown in Fig. \ref{fig:CMFig}. 
In these examples, our model provides more accurate results.
For instance, in the second row, BusterNet and CMSDNet can coarsely locate the forged regions without precise contours.  
The third sample has multiple copy-move objects. BusterNet and CMSDNet can not identify all the tampered objects in this challenging case. In contrast, our method not only locates all the manipulated areas, but also obtains sharp boundaries. 
In the last row of Fig. \ref{fig:CMFig}, only our method generates satisfactory localization results.

\subsection{Ablation Study}
In this subsection, we testify the necessity of some key components in the proposed TriPINet. 
In each ablation experiment, only one component is modified, while others remain unchanged. Then, we retrain the neural networks on CASIAv2 dataset with the same training protocol as before and report the results.   
\subsubsection{The Effectiveness of Different Backbones}  
\label{subsubsec:Backbone}
\begin{table}[t]
\centering
\renewcommand\arraystretch{1.5} 
\caption{\textbf{Ablation studies for backbones.} These backbones are initialized with the weights pre-trained on the ImageNet dataset. Then, our model with different backbones are trained and tested on CASIAv2 dataset.}
\label{tab:Ablation-Backbone}
\begin{tabular}{ccc}
\hline
\multirow{2}{*}{Backbone} & \multicolumn{2}{c}{CASIAv2}     \\ \cline{2-3} 
                          & F1             & IoU            \\ \hline
ResNet-18                 & 0.717          & 0.559          \\
ResNet-34                 & 0.737          & 0.584          \\
ResNet-50                 & 0.788          & 0.650          \\
VGG-11                    & 0.862          & 0.758          \\
\textbf{VGG-16}           & \textbf{0.904} & \textbf{0.824} \\ 
VGG-19                    & 0.807          & 0.677          \\ \hline
\end{tabular}
\end{table}

In this subsection, we evaluate our model with different backbones. These backbones are initialized with the weights that are pre-trained on the ImageNet dataset \cite{Russakovsky2015IJCV}.
As shown in Table. \ref{tab:Ablation-Backbone}, the performance of VGG networks are generally better than ResNet backbones. Since the VGG model is shallower than ResNet, the low-level features can supply more discriminative features for image manipulation localization \cite{Zhang2021TCSVT}. 
Our method achieves the best performance when using VGG-16 as backbones.  

\subsubsection{The Significance of the gCMDA module}
\begin{table}[t]
\centering
\renewcommand\arraystretch{1.5} 
\caption{\textbf{Ablation studies for the gCMDA module.} ``Freq-Noise" column denotes the fusion method for visual imperceptible featues. ``RGB" column indicates the operation for fusing RGB features with the visual imperceptible features.}
\label{tab:Ablation-gCMDA}
\begin{tabular}{ccccc}
\hline
\multicolumn{2}{c}{Feature Fusion} &  & \multicolumn{2}{c}{CASIAv2}     \\ \cline{1-2} \cline{4-5} 
Freq-Noise              & RGB            &  & F1             & IoU            \\ \hline
Add                     & Add            &  & 0.784          & 0.644          \\
Cat                     & Cat            &  & 0.705          & 0.545          \\
Add                     & GA             &  & 0.787          & 0.649          \\
Cat                     & GA             &  & 0.817          & 0.690          \\
CMDA                    & Add            &  & 0.800          & 0.667          \\
CMDA                    & Cat            &  & 0.710          & 0.550          \\
\textbf{CMDA}           & \textbf{GA}    &  & \textbf{0.862} & \textbf{0.758} \\ \hline
\end{tabular}
\end{table}
As shown in Table. \ref{tab:Ablation-gCMDA}, we conduct experiments to demonstrate the effectiveness of both CMDA and GA. 
In the first column of Table. \ref{tab:Ablation-gCMDA}, ``Freq-Noise" stands for how to fuse the features from frequency stream and noise stream. 
``RGB" in the second column represents how to fuse visual imperceptible features with RGB information. 
``Add" denotes elementwise addition, and ``Cat" indicates concatenate operation. 
It is obvious that the proposed gCMDA module can extract more discriminative cross-modality features for tampering localization. 
The reason can be interpreted as below.
The dual attention structure in CMDA module emphasizes the important features, while suppressing the redundant information both in channel and spatial dimensions. 
Moreover, the GA module uses RGB features as complementary semantic information. Since image manipulation aims to change the object layouts in the images, semantic clues is also useful in this task \cite{Dong2022TPAMI}.
In the second column of Fig. \ref{fig:AblationFig}, we illustrate some localization examples without gCMDA modules.
Intuitively, our model can not obtain satisfactory results without the gCMDA module. In this circumstance, hybrid features are not effectively fused.
  
\subsubsection{The Effectiveness of the PI-SE} 

\begin{table}[t]
\centering
\renewcommand\arraystretch{1.5} 
\caption{\textbf{Ablation studies for the PI-SE module.} ``3GAP+2GMP" indicates the decoder consists of three PI-SE modules with GAP in high-level stages, and two PI-SE modules with GMP in low-level stages.}
\label{tab:Ablation-PISE}
\begin{tabular}{ccc}
\hline
\multirow{2}{*}{PI-SE} & \multicolumn{2}{c}{CASIAv2}     \\ \cline{2-3} 
                       & F1             & IOU            \\ \hline
5GAP (w/o PI-SE)       & 0.797          & 0.663          \\
4GAP+1GMP              & 0.816          & 0.689          \\
\textbf{3GAP+2GMP }    & \textbf{0.862} & \textbf{0.758} \\
2GAP+3GMP              & 0.821          & 0.698          \\
1GAP+4GMP              & 0.797          & 0.676          \\
5GMP                   & 0.800          & 0.682          \\
4GMP+1GAP              & 0.816          & 0.698          \\
3GMP+2GAP              & 0.774          & 0.631          \\
2GMP+3GAP              & 0.789          & 0.651          \\
1GMP+4GAP              & 0.799          & 0.665          \\ \hline
\end{tabular}
\end{table}

In pixel-level visual tasks, how to gradually increase the feature maps in the decoder is vital for the final results \cite{LAP,Mao2021TMM,Chen2018CVPR}. To validate the proposed PI-SE modules, we alternately change the pooling operations, drawing the following observations. 
(1) In the first row of Table. \ref{tab:Ablation-PISE}, ``5GAP (w/o PI-SE)" denotes that all the PI-SE modules are replaced by the SE module \cite{Hu2018CVPR}.
Compared with conventional SE modules, hybrid pooling strategies bring obvious performance improvements in most cases. The advantage of GAP and GMP are fully explored by PI-SE modules in different stages.
(2) GAP at high-level stages and GMP at low-level stages perform better than the opposite order. 
In the high-level stages, each pixel is corresponding to a large receptive field, which contains significant information. 
We use PI-SE with GAP to recover the manipulated regions. 
In the low-level stages, the feature maps are restored by previous stages in the decoder. 
Thus, PI-SE with GMP is adopted to focus on the prominent information, ignoring abundant features such as false alarm areas. As a result, the boundaries of the predicted feature maps can be refined in detail \cite{Mao2021TMM}. 
In the fourth column of Fig. \ref{fig:AblationFig}, ``w/o PI-SE" represents for replacing PI-SE with conventional SE module. The proposed PI-SE module is helpful to precisely recover the feature maps.

\subsubsection{The Superiority of the Hybrid Loss}

\begin{table}[t]
\centering
\renewcommand\arraystretch{1.5} 
\caption{\textbf{Ablation studies for the hybrid loss function.}}
\label{tab:Ablation-Loss}
\begin{tabular}{ccc}
\hline
\multirow{2}{*}{Loss} & \multicolumn{2}{c}{CASIAv2}     \\ \cline{2-3} 
                      & F1             & IOU            \\ \hline
BCE                   & 0.640          & 0.470          \\
BCE+SSIM              & 0.631          & 0.461          \\
BCE+IOU               & 0.843          & 0.729          \\
\textbf{BCE+IOU+SSIM} & \textbf{0.862} & \textbf{0.758} \\ \hline
\end{tabular}
\end{table}

\begin{figure}[t]
	\centering
	\includegraphics[scale=0.1]{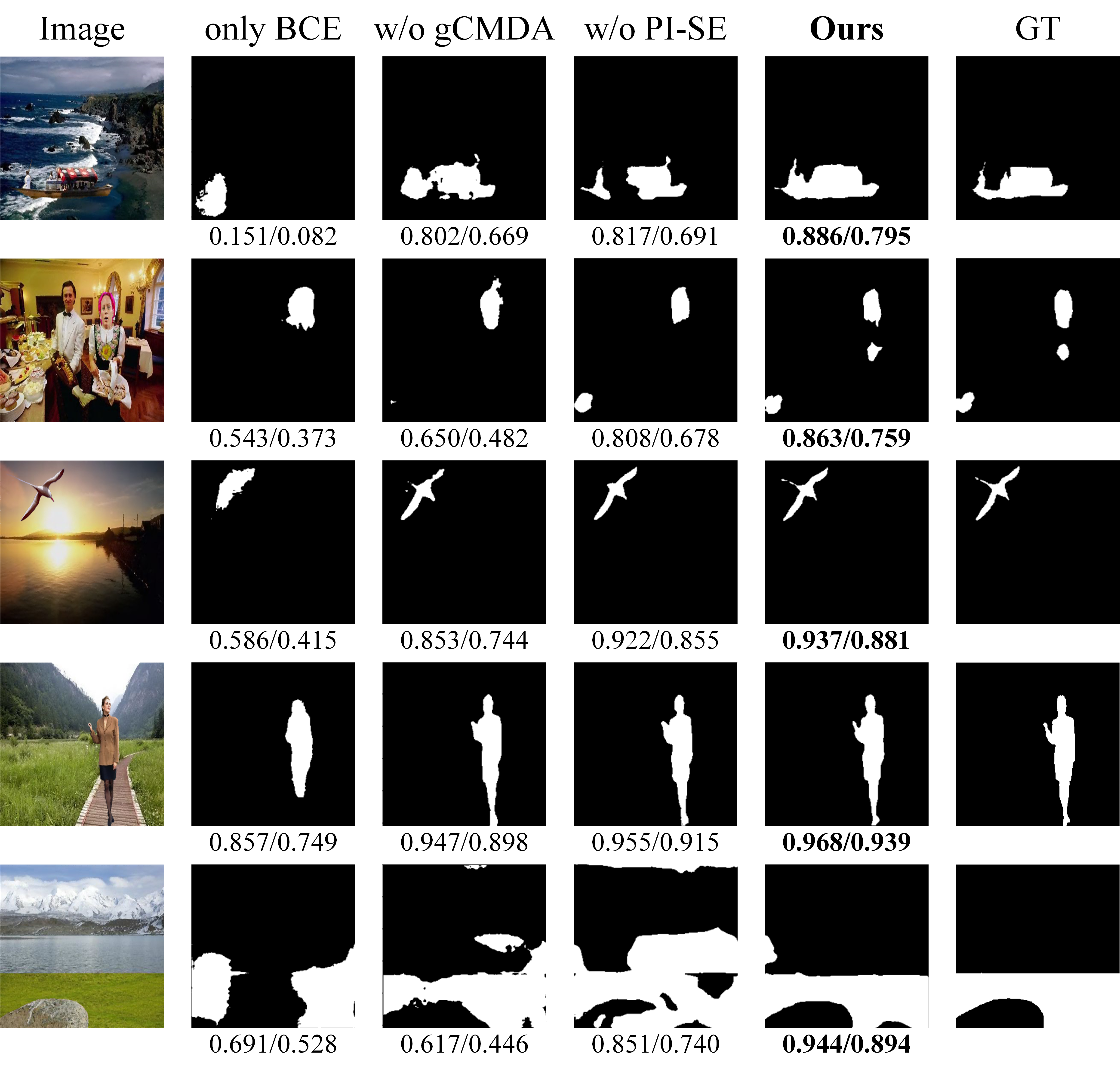}
	\centering
	\caption{\textbf{Ablation analysis of the proposed TriPINet (F1-score/IoU).} ``only BCE" denotes the model is only supervised by BCE loss. ``w/o gCMDA" indicates the model uses simple addition operations to replace the gCMDA module. ``w/o PI-SE" means that all the PI-SE modules are substituted with the SE module \cite{Hu2018CVPR}.}
	\label{fig:AblationFig}
\end{figure}

In Table. \ref{tab:Ablation-Loss}, we investigate the influence of hybrid loss. Comparing with simple BCE loss, the hybrid loss obtains a improvement by 22.2\%/25.9\% in terms of F1-score/IoU, which demonstrate the superiority of the hybrid loss. 
In the second column of Fig. \ref{fig:AblationFig}, the results supervised with BCE loss can only provide coarse shape of the manipulated regions. This demonstrates the hybrid loss can supervise the training process more effectively.

\subsection{Robustness Analysis}
\begin{figure}[t]
	\centering
	\subfigure[Against JPEG compression.]{
		\begin{minipage}[t]{0.5\linewidth}
			\centering
			\includegraphics[scale=0.25]{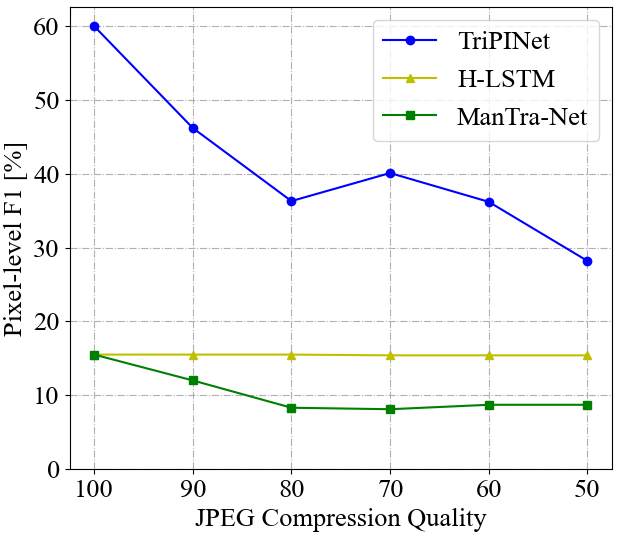}
			\label{fig:Robustness-JPEG}
		\end{minipage}
	}%
	\subfigure[Against Gaussian Blurs.]{
		\begin{minipage}[t]{0.5\linewidth}
			\centering
			\includegraphics[scale=0.25]{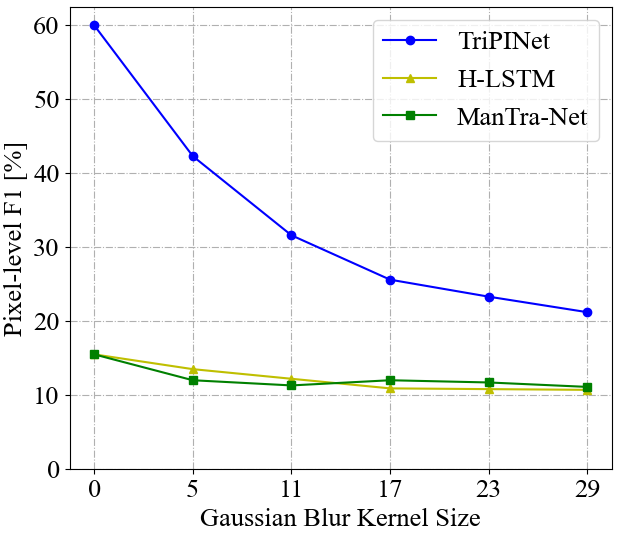}
			\label{fig:Robustness-GB}
		\end{minipage}
	}%
	\centering
	\caption{\textbf{Robustness evaluation against two image post-processing operations, e.g., JPEG Compression and Gaussian Blur.}}
	\label{fig:Robustness}
\end{figure}
We analyze the robustness of TriPINet against two common post-processing techniques when images transmitted on the Internet \cite{Zhou2018CVPR,Wu2019CVPR,Hu2020ECCV,Dong2022TPAMI}, i.e. JPEG compression and Gaussian blur. Following the robustness evaluation protocol defined in \cite{Dong2022TPAMI}, CASIAv2 is used for training, and CASIAv1 is adopted for test. Both the splicing and the copy-move forgeries are included. H-LSTM \cite{Bappy2019TIP_H-LSTM} and ManTra-Net \cite{Wu2019CVPR} are employed for comparison.

\subsubsection{Robustness Against JPEG Compression}
In the experiments, JPEG quality factors are ranged from 50 to 100, with the step of 10. 
Fig. \ref{fig:Robustness-JPEG} demonstrates that our method obtains the best localization results among the mentioned methods. 
When JPEG quality factor is below 70, image visual degradation and blocking artifact are noticeably visible \cite{Tralic2013CoMoFoD}. 
Our method can still correctly locate some forgeries in this challenging case.

\subsubsection{Robustness Against Gaussian Blurs}
In this subsection, we use five sizes of blur kernels, including $5\times5$, $11\times11$, $17\times17$, $23\times23$ and $29\times29$. In Fig. \ref{fig:Robustness-GB}, the proposed TriPINet obtains the best localization results among the compared methods. As the kernel size increased, the F1-score of TriPINet drops. The reason is inferred that the statistic distribution of blurred images is obviously different from the original ones, especially when the kernel size is bigger than $7 \times 7$ \cite{Tralic2013CoMoFoD}.

\subsection{Failure Cases and Analysis}   
\begin{figure}[t]
	\centering
	\includegraphics[scale=0.1]{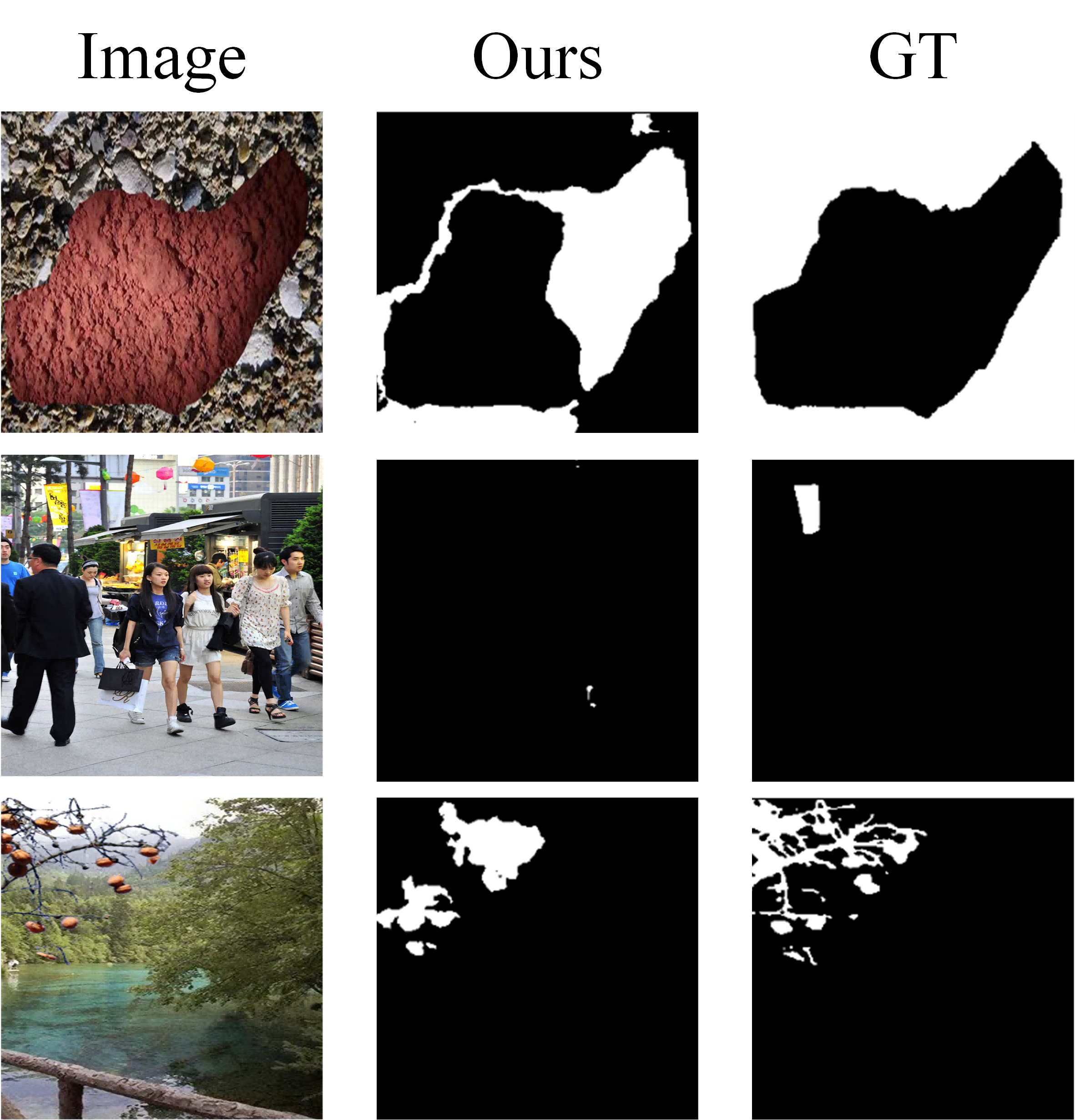}
	\centering
	\caption{Some failure cases of our proposed TriPINet.}
	\label{fig:FAIL}
\end{figure}
It is worthwhile to study the reasons for failure cases. The unsatisfactory results are mainly caused by three circumstances. First, the tampered regions are the background, occupying large areas. Our method will wrongly distinguish foreground and background in such situation, as demonstrated in the first row of Fig. \ref{fig:FAIL}. Second, when the foreground is small and the background contains complex scenes, the proposed TriPINet can not recognize tampered pixels well, such as the second row in Fig. \ref{fig:FAIL}. Finally, if the tampered object is complex with fine structures, our framework can not generate all the specifics of manipulated areas. This case is illustrated in the third row of Fig. \ref{fig:FAIL}. The tampered object is a tree with intricate details, such as lots of thin branches and small fruits.

\section{Conclusion}
In this paper, we propose a tripartite progressive integration network (TriPINet) for end-to-end image tampering localization. The gCMDA module is designed to fuse the complementary information both in visual perceptible and visual imperceptible domains. The PI-SE module is developed in the decoder stages to appropriately incorporate multiscale features and gradually recover the feature maps in detail. Our TriPINet is compared with 13 state-of-the-art methods on seven benchmark datasets, achieving comparable performance both in splicing and copy-move localization tasks. Ablation studies demonstrate the effectiveness of the proposed modules. Experimental results under attacks, e.g., JPEG compression and Gaussian blur, also show the robustness of the proposed TriPINet. Finally, some failure cases are discussed for further improvements in the future.

%

\bibliographystyle{IEEEtran}
\bibliography{TriPINet}

\begin{thebibliography}{10}
\providecommand{\url}[1]{#1}
\csname url@samestyle\endcsname
\providecommand{\newblock}{\relax}
\providecommand{\bibinfo}[2]{#2}
\providecommand{\BIBentrySTDinterwordspacing}{\spaceskip=0pt\relax}
\providecommand{\BIBentryALTinterwordstretchfactor}{4}
\providecommand{\BIBentryALTinterwordspacing}{\spaceskip=\fontdimen2\font plus
\BIBentryALTinterwordstretchfactor\fontdimen3\font minus
  \fontdimen4\font\relax}
\providecommand{\BIBforeignlanguage}[2]{{%
\expandafter\ifx\csname l@#1\endcsname\relax
\typeout{** WARNING: IEEEtran.bst: No hyphenation pattern has been}%
\typeout{** loaded for the language `#1'. Using the pattern for}%
\typeout{** the default language instead.}%
\else
\language=\csname l@#1\endcsname
\fi
#2}}
\providecommand{\BIBdecl}{\relax}
\BIBdecl

\bibitem{Verdoliva2020JSTSP}
L.~Verdoliva, ``{Media Forensics and DeepFakes: An Overview},'' \emph{IEEE
  Journal of Selected Topics in Signal Processing}, vol.~14, no.~5, pp.
  910--932, 2020.

\bibitem{Farid2019ARVS}
H.~Farid, ``Image forensics,'' \emph{Annual Review of Vision Science}, vol.~5,
  pp. 549--573, 2019.

\bibitem{Kaissis2020NMI}
G.~A. Kaissis, M.~R. Makowski, D.~R{\"u}ckert, and R.~F. Braren, ``Secure,
  privacy-preserving and federated machine learning in medical imaging,''
  \emph{Nature Machine Intelligence}, vol.~2, no.~6, pp. 305--311, 2020.

\bibitem{Han2021TBIOM}
B.~Han, X.~Han, H.~Zhang, J.~Li, and X.~Cao, ``{Fighting Fake News: Two Stream
  Network for Deepfake Detection via Learnable SRM},'' \emph{IEEE Transactions
  on Biometrics, Behavior, and Identity Science}, vol.~3, no.~3, pp. 320--331,
  2021.

\bibitem{Dong2022TPAMI}
C.~Dong, X.~Chen, R.~Hu, J.~Cao, and X.~Li, ``{MVSS-Net: Multi-View Multi-Scale
  Supervised Networks for Image Manipulation Detection},'' \emph{IEEE
  Transactions on Pattern Analysis and Machine Intelligence}, pp. 1--14, 2022.

\bibitem{Li2021TNNLS}
S.~Li, S.~Xu, W.~Ma, and Q.~Zong, ``{Image Manipulation Localization Using
  Attentional Cross-Domain CNN Features},'' \emph{IEEE Transactions on Neural
  Networks and Learning Systems}, pp. 1--15, 2021.

\bibitem{Wu2019CVPR}
Y.~Wu, W.~AbdAlmageed, and P.~Natarajan, ``{ManTra-Net: Manipulation tracing
  network for detection and localization of image forgeries with anomalous
  features},'' in \emph{Proceedings of the IEEE Conference on Computer Vision
  and Pattern Recognition}, 2019, pp. 9543--9552.

\bibitem{Hu2020ECCV}
X.~Hu, Z.~Zhang, Z.~Jiang, S.~Chaudhuri, Z.~Yang, and R.~Nevatia, ``{SPAN:
  Spatial pyramid attention network for image manipulation localization},'' in
  \emph{Proceedings of the European Conference on Computer Vision}, 2020, pp.
  312--328.

\bibitem{Rao2021ICCV}
Y.~Rao and J.~Ni, ``{Self-supervised domain adaptation for forgery localization
  of JPEG compressed images},'' in \emph{Proceedings of the IEEE International
  Conference on Computer Vision}, 2021, pp. 15\,034--15\,043.

\bibitem{Zhou2018CVPR}
P.~Zhou, X.~Han, V.~I. Morariu, and L.~S. Davis, ``{Learning Rich Features for
  Image Manipulation Detection},'' in \emph{Proceedings of the IEEE Conference
  on Computer Vision and Pattern Recognition}, 2018, pp. 1053--1061.

\bibitem{Ji2021IS}
Y.~Ji, H.~Zhang, Z.~Zhang, and M.~Liu, ``{CNN-based encoder-decoder networks
  for salient object detection: A comprehensive review and recent advances},''
  \emph{Information Sciences}, vol. 546, pp. 835--857, 2021.

\bibitem{Zhuo2022TIFS}
L.~Zhuo, S.~Tan, B.~Li, and J.~Huang, ``{Self-Adversarial Training
  Incorporating Forgery Attention for Image Forgery Localization},'' \emph{IEEE
  Transactions on Information Forensics and Security}, vol.~17, pp. 819--834,
  2022.

\bibitem{Zhuang2021TIFS}
P.~Zhuang, H.~Li, S.~Tan, B.~Li, and J.~Huang, ``Image tampering localization
  using a dense fully convolutional network,'' \emph{IEEE Transactions on
  Information Forensics and Security}, vol.~16, pp. 2986--2999, 2021.

\bibitem{Rao2021SP}
Y.~Rao, J.~Ni, and H.~Xie, ``{Multi-semantic CRF-based attention model for
  image forgery detection and localization},'' \emph{Signal Processing}, vol.
  183, p. 108051, 2021.

\bibitem{Bi2021ICCV_RTAG}
X.~Bi, Z.~Zhang, and B.~Xiao, ``{Reality Transform Adversarial Generators for
  Image Splicing Forgery Detection and Localization},'' in \emph{Proceedings of
  the IEEE International Conference on Computer Vision}, 2021, pp.
  14\,294--14\,303.

\bibitem{Wang2022PR}
X.~Wang, Y.~Wang, J.~Lei, B.~Li, Q.~Wang, and J.~Xue, ``Coarse-to-fine-grained
  method for image splicing region detection,'' \emph{Pattern Recognition},
  vol. 122, p. 108347, 2022.

\bibitem{Niu2021TIFS}
Y.~Niu, B.~Tondi, Y.~Zhao, R.~Ni, and M.~Barni, ``{Image Splicing Detection,
  Localization and Attribution via JPEG Primary Quantization Matrix Estimation
  and Clustering},'' \emph{IEEE Transactions on Information Forensics and
  Security}, vol.~16, pp. 5397--5412, 2021.

\bibitem{Liu2022TIP}
Y.~Liu, C.~Xia, X.~Zhu, and S.~Xu, ``{Two-Stage Copy-Move Forgery Detection
  With Self Deep Matching and Proposal SuperGlue},'' \emph{IEEE Transactions on
  Image Processing}, vol.~31, pp. 541--555, 2022.

\bibitem{Zhu2020TII_AR-Net}
Y.~Zhu, C.~Chen, G.~Yan, Y.~Guo, and Y.~Dong, ``{AR-Net: Adaptive Attention and
  Residual Refinement Network for Copy-Move Forgery Detection},'' \emph{IEEE
  Transactions on Industrial Informatics}, vol.~16, no.~10, pp. 6714--6723,
  2020.

\bibitem{Islam2020CVPR}
A.~Islam, C.~Long, A.~Basharat, and A.~Hoogs, ``{DOA-GAN: Dual-order attentive
  generative adversarial network for image copy-move forgery detection and
  localization},'' in \emph{Proceedings of the IEEE Conference on Computer
  Vision and Pattern Recognition}, 2020, pp. 4676--4685.

\bibitem{Wang2022CVPR}
J.~Wang, Z.~Wu, J.~Chen, X.~Han, A.~Shrivastava, S.-N. Lim, and Y.-G. Jiang,
  ``Objectformer for image manipulation detection and localization,'' in
  \emph{Proceedings of the IEEE Conference on Computer Vision and Pattern
  Recognition}, 2022, pp. 2364--2373.

\bibitem{Kwon2022IJCV}
M.-J. Kwon, S.-H. Nam, I.-J. Yu, H.-K. Lee, and C.~Kim, ``{Learning JPEG
  Compression Artifacts for Image Manipulation Detection and Localization},''
  \emph{International Journal of Computer Vision}, vol. 130, p. 1875–1895,
  2022.

\bibitem{Bappy2019TIP_H-LSTM}
J.~H. Bappy, C.~Simons, L.~Nataraj, B.~S. Manjunath, and A.~K. Roy-Chowdhury,
  ``{Hybrid LSTM and Encoder–Decoder Architecture for Detection of Image
  Forgeries},'' \emph{IEEE Transactions on Image Processing}, vol.~28, no.~7,
  pp. 3286--3300, 2019.

\bibitem{Hu2018CVPR}
J.~Hu, L.~Shen, and G.~Sun, ``{Squeeze-and-Excitation Networks},'' in
  \emph{Proceedings of the IEEE Conference on Computer Vision and Pattern
  Recognition}, 2018, pp. 7132--7141.

\bibitem{scSE}
A.~G. Roy, N.~Navab, and C.~Wachinger, ``Concurrent spatial and channel
  ‘squeeze \& excitation’ in fully convolutional networks,'' in
  \emph{Proceedings of the International Conference on Medical Image Computing
  and Computer-assisted Intervention}, 2018, pp. 421--429.

\bibitem{CBAM}
S.~Woo, J.~Park, J.-Y. Lee, and I.~S. Kweon, ``{CBAM: Convolutional block
  attention module},'' in \emph{Proceedings of the European Conference on
  Computer Vision}, 2018, pp. 3--19.

\bibitem{BAM}
J.~Park, S.~Woo, J.-Y. Lee, and I.~S. Kweon, ``{BAM: Bottleneck attention
  module},'' \emph{arXiv preprint arXiv:1807.06514}, 2018.

\bibitem{Gao2019CVPR}
Z.~Gao, J.~Xie, Q.~Wang, and P.~Li, ``Global second-order pooling convolutional
  networks,'' in \emph{Proceedings of the IEEE Conference on Computer Vision
  and Pattern Recognition}, 2019, pp. 3024--3033.

\bibitem{LAP}
Y.~Li, Q.~Miao, W.~Ouyang, Z.~Ma, H.~Fang, C.~Dong, and Y.~Quan, ``{LAP-Net:
  Level-aware progressive network for image dehazing},'' in \emph{Proceedings
  of the IEEE International Conference on Computer Vision}, 2019, pp.
  3276--3285.

\bibitem{Mao2021TMM}
Y.~Mao, Q.~Jiang, R.~Cong, W.~Gao, F.~Shao, and S.~Kwong, ``{Cross-Modality
  Fusion and Progressive Integration Network for Saliency Prediction on
  Stereoscopic 3D Images},'' \emph{IEEE Transactions on Multimedia}, vol.~24,
  pp. 2435--2448, 2021.

\bibitem{Chen2018CVPR}
H.~Chen and Y.~Li, ``{Progressively complementarity-aware fusion network for
  RGB-D salient object detection},'' in \emph{Proceedings of the IEEE
  Conference on Computer Vision and Pattern Recognition}, 2018, pp. 3051--3060.

\bibitem{Liu2022TCSVT}
X.~Liu, Y.~Liu, J.~Chen, and X.~Liu, ``{PSCC-Net: Progressive Spatio-Channel
  Correlation Network for Image Manipulation Detection and Localization},''
  \emph{IEEE Transactions on Circuits and Systems for Video Technology},
  vol.~32, no.~11, pp. 7505--7517, 2022.

\bibitem{Sun2022SPL}
Y.~Sun, R.~Ni, and Y.~Zhao, ``{ET: Edge-Enhanced Transformer for Image Splicing
  Detection},'' \emph{IEEE Signal Processing Letters}, vol.~29, pp. 1232--1236,
  2022.

\bibitem{Lee2015AISTATS}
C.-Y. Lee, S.~Xie, P.~Gallagher, Z.~Zhang, and Z.~Tu, ``{Deeply-Supervised
  Nets},'' in \emph{Proceedings of International Conference on Artificial
  Intelligence and Statistics}, vol.~38, 2015, pp. 562--570.

\bibitem{Qin2019CVPR}
X.~Qin, Z.~Zhang, C.~Huang, C.~Gao, M.~Dehghan, and M.~Jagersand, ``{BASNet:
  Boundary-Aware Salient Object Detection},'' in \emph{Proceedings of the IEEE
  Conference on Computer Vision and Pattern Recognition}, 2019, pp. 7479--7489.

\bibitem{BCE}
P.-T. De~Boer, D.~P. Kroese, S.~Mannor, and R.~Y. Rubinstein, ``A tutorial on
  the cross-entropy method,'' \emph{Annals of Operations Research}, vol. 134,
  no.~1, pp. 19--67, 2005.

\bibitem{Simonyan2014VGG}
K.~Simonyan and A.~Zisserman, ``{Very Deep Convolutional Networks for
  Large-scale Image Recognition},'' \emph{arXiv preprint arXiv:1409.1556},
  2014.

\bibitem{Russakovsky2015IJCV}
O.~Russakovsky, J.~Deng, H.~Su, J.~Krause, S.~Satheesh, S.~Ma, Z.~Huang,
  A.~Karpathy, A.~Khosla, M.~Bernstein \emph{et~al.}, ``{Imagenet Large Scale
  Visual Recognition Challenge},'' \emph{International Journal of Computer
  Vision}, vol. 115, no.~3, pp. 211--252, 2015.

\bibitem{Gao2022TKDE_TBNet}
Z.~Gao, C.~Sun, Z.~Cheng, W.~Guan, A.~Liu, and M.~Wang, ``{TBNet: A Two-Stream
  Boundary-Aware Network for Generic Image Manipulation Localization},''
  \emph{IEEE Transactions on Knowledge and Data Engineering}, pp. 1--16, 2022.

\bibitem{Ahmed1974DCT}
N.~Ahmed, T.~Natarajan, and K.~R. Rao, ``{Discrete Cosine Transform},''
  \emph{IEEE Transactions on Computers}, vol. 100, no.~1, pp. 90--93, 1974.

\bibitem{Bayar2018TIFS}
B.~Bayar and M.~C. Stamm, ``{Constrained Convolutional Neural Networks: A New
  Approach Towards General Purpose Image Manipulation Detection},'' \emph{IEEE
  Transactions on Information Forensics and Security}, vol.~13, no.~11, pp.
  2691--2706, 2018.

\bibitem{Fridrich2012TIFS}
J.~Fridrich and J.~Kodovsky, ``{Rich Models for Steganalysis of Digital
  Images},'' \emph{IEEE Transactions on Information Forensics and Security},
  vol.~7, no.~3, pp. 868--882, 2012.

\bibitem{Li2019ICCV}
H.~Li, G.~Chen, G.~Li, and Y.~Yu, ``{Motion Guided Attention for Video Salient
  Object Detection},'' in \emph{Proceedings of the IEEE International
  Conference on Computer Vision}, 2019, pp. 7274--7283.

\bibitem{Cao2022CVPR}
J.~Cao, C.~Ma, T.~Yao, S.~Chen, S.~Ding, and X.~Yang, ``{End-to-End
  Reconstruction-Classification Learning for Face Forgery Detection},'' in
  \emph{Proceedings of the IEEE Conference on Computer Vision and Pattern
  Recognition}, 2022, pp. 4113--4122.

\bibitem{Zhang2021TCSVT}
Y.~Zhang, G.~Zhu, L.~Wu, S.~Kwong, H.~Zhang, and Y.~Zhou, ``{Multi-task
  SE-Network for Image Splicing Localization},'' \emph{IEEE Transactions on
  Circuits and Systems for Video Technology}, vol.~32, no.~7, pp. 4828--4840,
  2022.

\bibitem{Wu2022TCSVT}
H.~Wu and J.~Zhou, ``{IID-Net: Image Inpainting Detection Network via Neural
  Architecture Search and Attention},'' \emph{IEEE Transactions on Circuits and
  Systems for Video Technology}, vol.~32, no.~3, pp. 1172--1185, 2022.

\bibitem{Shore1980TIT}
J.~Shore and R.~Johnson, ``{Axiomatic Derivation of the Principle of Maximum
  Entropy and the Principle of Minimum Cross-entropy},'' \emph{IEEE
  Transactions on Information Theory}, vol.~26, no.~1, pp. 26--37, 1980.

\bibitem{Qin2021arXiv}
X.~Qin, D.-P. Fan, C.~Huang, C.~Diagne, Z.~Zhang, A.~C. Sant'Anna, A.~Suarez,
  M.~Jagersand, and L.~Shao, ``{Boundary-aware segmentation network for mobile
  and web applications},'' \emph{arXiv preprint arXiv:2101.04704}, 2021.

\bibitem{Yu2016ACMMM}
J.~Yu, Y.~Jiang, Z.~Wang, Z.~Cao, and T.~Huang, ``{Unitbox: An Advanced Object
  Detection Network},'' in \emph{Proceedings of the ACM International
  Conference on Multimedia}, 2016, pp. 516--520.

\bibitem{Wang2004TIP}
Z.~Wang, A.~C. Bovik, H.~R. Sheikh, and E.~P. Simoncelli, ``{Image Quality
  Assessment: from Error Visibility to Structural Similarity},'' \emph{IEEE
  Transactions on Image Processing}, vol.~13, no.~4, pp. 600--612, 2004.

\bibitem{Dong2013CASIA}
J.~Dong, W.~Wang, and T.~Tan, ``{CASIA Image Tampering Detection Evaluation
  Database},'' in \emph{Proceedings of IEEE China Summit and International
  Conference on Signal and Information Processing}, 2013, pp. 422--426.

\bibitem{Ng2009COLUMB}
T.-T. Ng, J.~Hsu, and S.-F. Chang, ``{Columbia Image Splicing Detection
  Evaluation Dataset},'' \emph{DVMM lab. Columbia Univ CalPhotos Digit Libr},
  2009.

\bibitem{De2013Carvalho}
T.~J. De~Carvalho, C.~Riess, E.~Angelopoulou, H.~Pedrini, and
  A.~de~Rezende~Rocha, ``{Exposing Digital Image Forgeries by Illumination
  Color Classification},'' \emph{IEEE Transactions on Information Forensics and
  Security}, vol.~8, no.~7, pp. 1182--1194, 2013.

\bibitem{Wu2018ECCV_BusterNet}
Y.~Wu, W.~Abd-Almageed, and P.~Natarajan, ``{BusterNet: Detecting copy-move
  image forgery with source/target localization},'' in \emph{Proceedings of the
  European Conference on Computer Vision}, 2018, pp. 168--184.

\bibitem{Tralic2013CoMoFoD}
D.~Tralic, I.~Zupancic, S.~Grgic, and M.~Grgic, ``{CoMoFoD-New Database For
  Copy-Move Forgery Detection},'' in \emph{Proceedings of ELMAR}, 2013, pp.
  49--54.

\bibitem{Wen2016COVERAGE}
B.~Wen, Y.~Zhu, R.~Subramanian, T.-T. Ng, X.~Shen, and S.~Winkler,
  ``{COVERAGE-A Novel Database for Copy-Move Forgery Detection},'' in
  \emph{Proceedings of IEEE International Conference on Image Processing},
  2016, pp. 161--165.

\bibitem{Salloum2018JVCIR_MFCN}
R.~Salloum, Y.~Ren, and C.-C.~J. Kuo, ``{Image Splicing Localization using a
  Multi-task Fully Convolutional Network (MFCN)},'' \emph{Journal of Visual
  Communication and Image Representation}, vol.~51, pp. 201--209, 2018.

\bibitem{Bappy2017ICCV_J-LSTM}
J.~H. Bappy, A.~K. Roy-Chowdhury, J.~Bunk, L.~Nataraj, and B.~Manjunath,
  ``{Exploiting Spatial Structure for Localizing Manipulated Image Regions},''
  in \emph{Proceedings of the IEEE International Conference on Computer
  Vision}, 2017, pp. 4970--4979.

\bibitem{Cun2018ECCVW}
X.~Cun and C.-M. Pun, ``{Image Splicing Localization via Semi-Global Network
  and Fully Connected Conditional Random Fields},'' in \emph{Proceedings of the
  European Conference on Computer Vision Workshops}, 2018, pp. 1--15.

\bibitem{Bi2019CVPRW_RRU-Net}
X.~Bi, Y.~Wei, B.~Xiao, and W.~Li, ``{RRU-Net: The Ringed Residual U-Net for
  Image Splicing Forgery Detection},'' in \emph{Proceedings of the IEEE
  Conference on Computer Vision and Pattern Recognition Workshops}, 2019, pp.
  1--10.

\bibitem{Xiao2020IS_C2RNet}
B.~Xiao, Y.~Wei, X.~Bi, W.~Li, and J.~Ma, ``{Image Splicing Forgery Detection
  Combining Coarse to Refined Convolutional Neural Network and Adaptive
  Clustering},'' \emph{Information Sciences}, vol. 511, pp. 172--191, 2020.

\bibitem{Wei2020TOMM}
Y.~Wei, Z.~Wang, B.~Xiao, X.~Liu, Z.~Yan, and J.~Ma, ``{Controlling Neural
  Learning Network with Multiple Scales for Image Splicing Forgery
  Detection},'' \emph{ACM Transactions on Multimedia Computing, Communications,
  and Applications}, vol.~16, no.~4, pp. 1--22, 2020.

\bibitem{Chen2021TMM_CMSDNet}
B.~Chen, W.~Tan, G.~Coatrieux, Y.~Zheng, and Y.-Q. Shi, ``{A Serial Image
  Copy-Move Forgery Localization Scheme With Source/Target Distinguishment},''
  \emph{IEEE Transactions on Multimedia}, vol.~23, pp. 3506--3517, 2021.

\end{thebibliography}

\vline

\end{document}